\documentclass[runningheads]{llncs}

\usepackage{eccv}

\usepackage{eccvabbrv}

\usepackage{graphicx}
\usepackage{svg}
\usepackage{booktabs}
\usepackage{multirow}
\usepackage{makecell}
\usepackage{wrapfig}
\usepackage{tcolorbox}
\tcbuselibrary{breakable}
\makeatletter
\def\blfootnote{\gdef\@thefnmark{}\@footnotetext}
\makeatother

\usepackage[accsupp]{axessibility}  % Improves PDF readability for those with disabilities.

\usepackage{hyperref}

\usepackage{orcidlink}
\usepackage{todonotes}
\usepackage{listings}

\usepackage[table]{xcolor}
\definecolor{lightgreen}{HTML}{B6DEC2}
\definecolor{lightred}{HTML}{FCCAC5}
\definecolor{lightblue}{HTML}{C8EAF5}
\definecolor{Highlight}{HTML}{fc8d62}  % green

\newlength\savewidth

\usepackage{array}
\newcolumntype{x}[1]{>{\centering\arraybackslash}p{#1pt}}
\newcolumntype{y}[1]{>{\raggedright\arraybackslash}p{#1pt}}
\newcolumntype{z}[1]{>{\raggedleft\arraybackslash}p{#1pt}}

\newcommand{\methodname}{Prompt2Effect}
\definecolor{projectorange}{rgb}{0.90,0.49,0.13}
\begin{document}

\title{
% Prompt2Effect: Training-Free LoRA Synthesis for Instant Video Effects \\
% Prompt2Effect: On-the-Fly LoRA Prediction for Text-to-Video Effect Control \\ 
% Prompt2Effect: On-the-Fly LoRA Prediction for Instant Text-to-Video Update \\
% Prompt2Effect: On-the-Fly LoRA Prediction for Instant User Control in Text-to-Video \\
% Prompt2Effect: LoRA Generation for Training-Free Effect Personalization of Video Diffusion Models \\
Prompt2Effect: Training-Free Image-to-Video Model Specialization via LoRA Generation
}

\titlerunning{Prompt2Effect}

\author{Xiaomeng Yang\inst{1,2}\thanks{Work done during internship at Snap Inc.} \and
Yanyu Li \inst{1} \and
Gordon Guocheng Qian\inst{1} \and
Ivan Skorokhodov\inst{1} \and 
Viacheslav Ivanov\inst{1} \and
Avalon Vinella\inst{1} \and Xuan Zhang\inst{2} \and Yanzhi Wang\inst{2} \and Sergey Tulyakov\inst{1} \and Anil Kag\inst{1}
}

\authorrunning{X. Yang et al.}

\institute{$^1$Snap Inc.,  $^2$Northeastern University\\
\href{https://xiaomeng-yang.github.io/Prompt2Effect/}{\textcolor{projectorange}{https://xiaomeng-yang.github.io/Prompt2Effect/}}}
% \url{http://www.springer.com/gp/computer-science/lncs}}

% %%%%%%%%%%%%%%%
% % TEASER START
% %%%%%%%%%%%%%%%
\makeatletter
\let\@oldmaketitle\@maketitle% Store \@maketitle
\renewcommand{\@maketitle}{\@oldmaketitle% Update \@maketitle to insert...
\myfigure\bigskip}
\makeatother
\newcommand\myfigure{%
  \makebox[0pt]{\hspace{11.5cm}  \includegraphics[width=\linewidth]{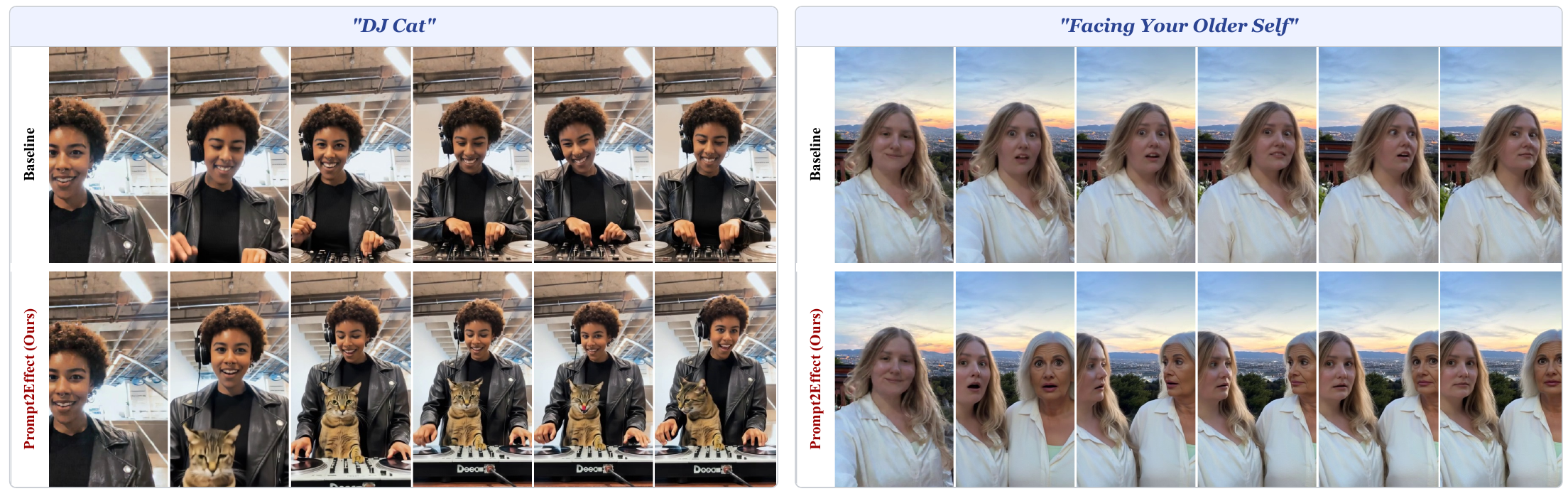}}
  \\
\refstepcounter{figure}\textbf{Fig. \thefigure}: \methodname{} enables zero-shot LoRA synthesis through a single forward pass for effect-injected image-to-video diffusion generation. 
\label{fig:teaser}
}
% \makeatother
% %%%%%%%%%%%%
% % TEASER END
% %%%%%%%%%%%%

\maketitle

\begin{abstract}
While personalizing Image-to-Video (I2V) diffusion models with specific visual effects is increasingly demanded for high-end generation, current practice requires training a separate Low-Rank Adaptation (LoRA) module for each effect, incurring substantial data curation and iterative optimization costs that hinder interactive control. We present \textit{Prompt2Effect}, a weight-driven hypernetwork that amortizes per-effect training by directly synthesizing effect-specific LoRA weights in a single forward pass. Unlike prior hypernetworks that regress adapter weights purely from semantics, \textit{Prompt2Effect} is explicitly conditioned on the frozen base model weights, grounding prediction in the structural geometry of each layer. Furthermore, instead of predicting raw LoRA matrices, we introduce an SVD-canonicalized parameterization that resolves factorization ambiguity and stabilizes large-scale synthesis. Extensive experiments demonstrate that Prompt2Effect achieves on-par or superior video quality and effect alignment compared to conventional LoRA fine-tuning, while reducing the computational cost from 56 GPU training hours to 3.3 seconds of hypernetwork inference. When used as initialization for subsequent fine-tuning, our predicted weights further improve final performance and accelerate optimization by approximately $10\times$.
\keywords{Video Diffusion, Video Effect, LoRA, Hypernetwork, Personalization}
\end{abstract}

% \textcolor{red}{Add Citations, }
% \textcolor{red}{Teasure and Method Figure} 

\section{Introduction} \label{sec:intro}
Recent Image-to-Video (I2V) diffusion models~\cite{wan2025wan} have achieved remarkable success in synthesizing high-fidelity video from an input image and text guidance. Beyond general semantic control, there is a burgeoning demand for effect-level personalization: the ability to render an input image into specific visual styles or motion effects with fine-grained, immediate control. In practice, these complex dynamic effects are often distilled from dozens of unstructured reference videos, which are far too heavy to use as direct conditions during inference. Consequently, the standard approach for such personalization~\cite{abdal2025dynamic,mao2025omni} involves training a separate Low-Rank Adaptation (LoRA)~\cite{hu2022lora} module on a new video dataset consisting of the target effects. However, per-effect data curation and LoRA training are inherently iterative and computationally expensive, often requiring thousands of optimization steps, which create a significant bottleneck for interactive creative workflows. This overhead is particularly acute for I2V models, where the increased model parameters and temporal complexity make effect authoring slow and difficult to scale~\cite{peebles2023scalable}.

To bypass the costs of gradient-based optimization and enable \emph{amortized per-effect LoRA training}, weight-generation networks such as hypernetworks~\cite{ha2016hypernetworks, charakorn2025text} have emerged as a compelling alternative. These models directly predict model parameters in a single forward pass, conditioned on a target task. This raises a natural question: can we leverage this paradigm to synthesize effect-specific LoRA weights for I2V generation models directly from a textual effect description, without per-effect optimization? For image generation, hypernetwork approaches such as HyperDreamBooth~\cite{ruiz2024hyperdreambooth} have already demonstrated the potential of training-free personalization by generating adapter weights conditioned on an input face image. 

However, directly extending these techniques to I2V diffusion remains highly non-trivial. First, existing methods primarily focus on localized \emph{subject/identity} injection in static images (typically using UNet architectures~\cite{rombach2022high}), whereas our goal is to control global dynamic \emph{visual effects and motion} in video. Second, modern I2V models (such as Diffusion Transformers~\cite{wan2025wan}) involve high-dimensional temporal latent dynamics and require consistent adaptation across an exponentially larger parameter space. In practice, existing weight generation designs face a fundamental scaling challenge: they struggle to regress the large and structured LoRA parameter space of I2V models and can fail to converge to usable weight predictors under this setting (\cref{fig:ablation_svd}).

To address these challenges, we propose \textbf{\methodname{}}, a \textbf{weight-driven hypernetwork for SVD-canonicalized LoRA prediction} in I2V diffusion models. 
Unlike prior hypernetwork approaches that rely solely on input conditions such as effect prompts or identity images, our hypernetwork is explicitly conditioned on the frozen base weights of the target model. 
This weight-driven formulation is motivated by the nature of LoRA itself: a LoRA update $\Delta W$ is not an absolute parameter, but a structured adaptation defined with respect to the base model weight matrix $W_0$. 
When the predictor has no access to $W_0$, it must implicitly infer the geometry of the base layer purely from semantics, which becomes increasingly ill-posed as model dimensionality grows. 
By conditioning on $W_0$, the hypernetwork is grounded in the native input-output coordinates and correlation structure of each layer, substantially reducing the ambiguity of large-scale weight synthesis.

Furthermore, existing methods typically regress LoRA matrices in their raw factorized form. 
However, the decomposition $\Delta W = BA$ is inherently non-identifiable: for any invertible matrix $R$, the pairs $(BR, R^{-1}A)$ produce the same update. 
This induces a highly redundant solution space and leads to unstable optimization when predicting many layer-wise adapters simultaneously. 
To resolve this ambiguity, we predict the SVD-canonicalized form of the LoRA update. 
By decomposing $\Delta W = U S V^\top$ and predicting the canonical factors $B^\star = U S^{1/2}$ and $A^\star = S^{1/2} V^\top$, we enforce an orthogonal, energy-ordered parameterization with a unique representation up to sign. 
This significantly constrains the prediction space, improves convergence, and stabilizes large-scale LoRA weight synthesis.
Together, these two design principles, weight-driven conditioning and SVD-canonicalized prediction, make \methodname{} possible to stably synthesize high-dimensional LoRA updates for modern I2V models in a single forward pass (see ablation in \cref{fig:ablation_svd}), enabling accurate training-free effect control (\cref{fig:teaser}).

Our key \textbf{contributions} are summarized as follows:
\vspace{-0.5em}
\begin{itemize}
\item \textbf{Weight-Driven LoRA Synthesis for I2V Diffusion.} 
We introduce a weight-conditioned formulation for LoRA prediction that explicitly leverages frozen base weights to ground adaptation in layer geometry, enabling stable scaling to high-dimensional I2V diffusion models.

\item \textbf{SVD-Canonicalized LoRA Prediction.} 
We identify the non-identifiability of weight regression as a fundamental obstacle to stable training. 
To address this, we propose predicting SVD-canonicalized LoRA factors, which impose an orthogonal and energy-ordered parameterization, reduce redundancy in the solution space, and significantly improve convergence stability.

\item \textbf{Scalable Training-Free Effect Control.} 
Our framework amortizes the training process, enabling instantaneous effect generation without per-effect optimization by synthesizing large collections of layer-wise LoRA updates in a single forward pass. \methodname{} achieves on-par or superior video quality and effect alignment compared to conventional LoRA fine-tuning, while reducing the computational cost from 56 GPU hours of training to 3.3 seconds of inference time per effect.

\item \textbf{Fast Adaptation and Composable Control.} 
When maximum fidelity is required, our predicted weights serve as strong initialization, substantially accelerating conventional LoRA fine-tuning by $10\times$. 
Moreover, \methodname{} supports interpolation and composition directly in weight space, enabling flexible and continuous effect manipulation.
\end{itemize}

\section{Related Work} \label{sec.related}

\noindent\textbf{Controllable Video Generation.}
While controllable video generation includes \emph{structure-controlled} methods leveraging pixel-aligned conditions, our work focuses on \emph{semantic/effect} control to enforce non-aligned target behaviors like concepts, styles, or VFX without pixel-wise correspondence.

\emph{Per-effect / per-video adaptation.}
A dominant approach to semantic control is \emph{condition-specific overfitting}, adapting a pretrained T2V model to each new effect via lightweight finetuning like DreamBooth~\cite{ruiz2023dreambooth} or LoRA~\cite{hu2022lora}. For instance, Set-and-Sequence~\cite{abdal2025dynamic} isolates identity and motion through two-stage LoRA fine-tuning, but still requires per-concept optimization. Similarly, models like VFX Creator~\cite{liu2025vfx}, Omni-Effects~\cite{mao2025omni}, and EasyVFX~\cite{ma2026easyvfx} train specialized adapters or effect experts, yet they rely on predefined categories or effect-specific training that cannot generalize to unseen effects without further adaptation.

\emph{Task-specific design and inference-time control.}
Another line of work improves controllability through task-specific modules or inference-time mechanisms. Tune-A-Video~\cite{wu2023tune} preserves structure while injecting new semantics via one-shot video adaptation with inversion-based initialization. For reference-driven stylization, StyleMaster~\cite{ye2025stylemaster} employs adapter-like cross-attention to condition generation on style images, similar to IP-Adapter~\cite{ye2023ip-adapter}. Recent works also steer generation without explicit parameter updates by augmenting inputs with additional modalities or latents~\cite{junhao2024moonshot,abdal2025zero,bian2025video}. While ControlNet-style branches and image-conditioned adapters excel with compact conditions like masks or static images, they struggle with our target application. Global dynamic effects are unstructured and typically require dozens of reference videos, making them too computationally heavy for inference-time conditioning. In contrast, our goal is \emph{zero-shot} effect control at inference time by \emph{synthesizing} plug-and-play LoRA parameters from an effect prompt in a single forward pass.

\noindent\textbf{Weight Prediction Networks.}
Hypernetworks~\cite{ha2016hypernetworks} that predict adaptation weights have been explored as an efficient alternative to per-instance finetuning. HyperDreamBooth~\cite{ruiz2024hyperdreambooth} predicts personalized diffusion-model weights for fast identity adaptation, demonstrating that weight generation can substantially reduce test-time cost. Subsequent works~\cite{lv2024hyperlora, charakorn2025text} improved parameter efficiency by predicting LoRA~\cite{hu2022lora} adapters rather than full weights. Notably, LoFA~\cite{hao2025lofa} serves as recent prior art in weight-driven LoRA prediction. However, LoFA targets compact conditions (e.g., style, pose reference images) and directly regresses ambiguous raw LoRA factors, which can lead to unstable optimization. In contrast, our framework synthesizes LoRAs for unstructured, dynamic effects distilled from dozens of videos, and we propose an SVD-canonicalized prediction to resolve factorization ambiguity and stabilize large-scale weight synthesis. 

More recently, generative approaches have been proposed to synthesize LoRA parameters, including diffusion-based LoRA generation for LLMs and image diffusion models~\cite{khan2025oral,liang2025draganddrop,wu2024difflora}. Despite this progress, existing methods primarily focus on LLMs or localized \emph{subject/identity} injection in static images (typically UNet architectures). Extending \emph{prompt-conditioned LoRA synthesis} to large-scale \emph{video} diffusion backbones (such as Diffusion Transformers~\cite{wan2025wan}) remains underexplored. This transition is highly non-trivial due to the explosion in parameter dimensionality and the need to adapt temporally structured representations for global visual effects.

Our \methodname{} addresses this challenge by introducing a hypernetwork that predicts SVD-canonicalized LoRA weights from effect prompts. While recent works like PiSSA~\cite{meng2024pissa} and Make-a-LoRA~\cite{fan2025make} have shown that SVD-based LoRA initialization improves gradient-based training convergence, we uniquely leverage SVD to canonicalize the regression target space for hypernetwork training. This ensures a structurally aligned and stable optimization landscape for the hypernetwork, allowing it to reliably predict complex video effect parameters.
\section{Method}
In this section, we introduce \methodname{}, a framework for training-free LoRA synthesis at inference time for controllable video effects. Our overall pipeline consists of two main phases: (1) a one-time pretraining stage, where a hypernetwork is trained to regress LoRA weights from effect prompts, and (2) a zero-shot inference stage, where the hypernetwork directly predicts effect-specific LoRA weights from a given effect prompt through a single forward pass. We also introduce an optional lightweight testing-time optimization that uses the predicted weights as an initialization to improve the zero-shot quality while being $10\times$ faster than standard LoRA training from scratch.

\subsection{Preliminaries}

\noindent\textbf{Video Diffusion Models.} Video diffusion models generate video clips by iteratively denoising a sequence of latent variables. Given a clean video latent $z_0$, a forward diffusion process adds Gaussian noise to produce a noisy latent $z_t$ at timestep $t$. The model, typically parameterized by a 3D-UNet~\cite{rombach2022high} or spatial-temporal Transformer~\cite{peebles2023scalable}, is trained to reverse the process and generate data $z_0$ with conditions such as textual prompts ($c$).

\noindent\textbf{Low-Rank Adaptation (LoRA).} LoRA~\cite{hu2022lora,ruiz2023dreambooth} is a parameter-efficient finetuning technique that adapts a frozen pretrained model to new concepts or domains. Instead of updating the original weight matrix $W \in \mathbb{R}^{d \times k}$, LoRA optimizes a low-rank residual $\Delta W$:
$$W' = W + \Delta W = W + BA$$
where $B \in \mathbb{R}^{d \times r}$ and $A \in \mathbb{R}^{r \times k}$ are trainable low-rank matrices, and the rank $r \ll \min(d, k)$.

\noindent\textbf{Hypernetworks for LoRA Prediction.} A hypernetwork $H_{\phi}$ parameterized by $\phi$ is a meta-model designed to generate the weights of another neural network. In our context, we aim to learn a mapping function that takes an effect prompt $c_{\text{eff}}$ as input and directly predicts a collection of LoRA weight updates for a pretrained video diffusion model:
\begin{equation}
\Delta \Theta = H_{\phi}(c_{\text{eff}}), \quad
\Delta \Theta = \{ \Delta W_i \}_{i=1}^{L}
\end{equation}
where $\Delta \Theta$ represents the set of low-rank adaptations applied to the $L$ selected layers 
of the base model with pretrained parameters $\Theta_0$.

The adapted model parameters are obtained via
\begin{equation}
\Theta = \Theta_0 + \Delta \Theta.
\end{equation}

Under the LoRA parameterization, each layer update is decomposed as

\begin{equation}
\Delta W^l = B^l A^l, 
\quad 
A^l \in \mathbb{R}^{r \times d_{in}^l}, 
\quad 
B^l \in \mathbb{R}^{d_{out}^l \times r}, 
\quad 
l = 1, \dots, L,
\end{equation}

\begin{figure}[t]
    \centering
\includegraphics[width=\linewidth]{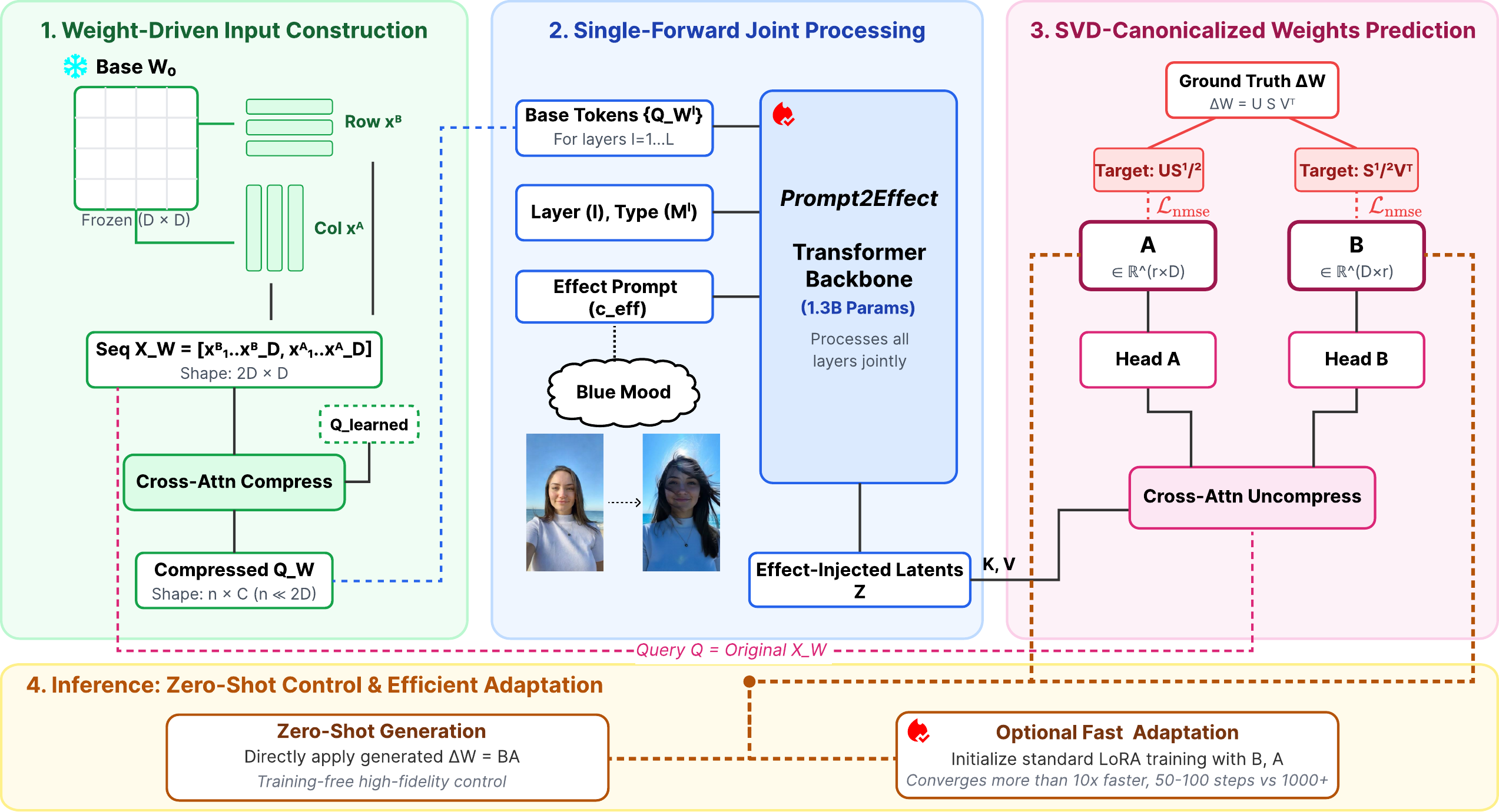}
\caption{Overview of the \methodname{} framework. The pipeline consists of three stages: 
(1) \textbf{Weight-Driven Input Construction}, where the frozen base weights $W_0$ are decomposed and compressed into compact weight tokens; 
(2) \textbf{Hypernetwork Backbone}, which jointly processes the compressed weight tokens, effect prompt, and layer metadata to produce effect-injected latents; and 
(3) \textbf{SVD-Canonicalized Weight Prediction}, which projects the latents back to the original parameter space to predict SVD-canonicalized LoRA matrices. 
The predicted weights can be directly applied for zero-shot generation, or used as initialization for fast adaptation on more complex effects to further improve performance with more than $10\times$ training reduction compared to LoRA training from scratch. Here we use $D$ for simplicity.}
    \label{fig:method}
\vspace{-1em}
\end{figure}
\subsection{Weight-Driven SVD LoRA Prediction}
Existing hypernetworks~\cite{wu2024difflora,ruiz2024hyperdreambooth} typically generate target network weights directly from abstract semantic embeddings, such as noise vectors or text features. 
These approaches are \emph{purely semantic-driven}: the predictor has no explicit access to the base model weights it is meant to adapt. 
As a result, weight generation lacks structural grounding in the underlying network, which empirically leads to suboptimal performance when predicting high-dimensional updates (Sec.~\ref{sec:ablation}).

Moreover, prior methods generally regress weights in their raw parameterization. 
Directly predicting unconstrained weight matrices introduces two additional challenges: 
(1) the solution space is highly non-identifiable due to equivalent factorizations of $\Delta W = BA$; thus, for any invertible matrix $R$, the pairs $(BR, R^{-1}A)$ produce the same update, and 
(2) the absence of canonical structure makes optimization less stable, especially for large networks.

To address these limitations, we propose a \textit{weight-driven SVD LoRA prediction} framework. 
First, instead of relying solely on semantic embeddings, we condition the hypernetwork explicitly on the frozen base weights $W_0$, providing strong structural priors for predicting $\Delta W$. 
Second, rather than regressing raw LoRA matrices, we predict their SVD-canonicalized form, which removes factorization ambiguity and stabilizes training.

\noindent\textbf{Weight-Driven Input Formulation.}
We formulate a novel \textit{weight-driven} prediction mechanism, motivated by the fact that a LoRA update is an adaptation defined with respect to the original weight matrix. Therefore, we propose to use the pre-trained network weights directly as an extra input along with the textual prompt to the hypernetwork to predict the desired LoRA update. 

Specifically, for a target layer $l$ with frozen baseline weights $W_0^l \in \mathbb{R}^{d^l_{\text{out}} \times d^l_{\text{in}}}$, we construct a structured input sequence by slicing $W^l_0$ into \emph{row tokens} and \emph{column tokens}. The rows form \emph{B tokens} $x^{B,l}_i = W^l_0[i,:] \in \mathbb{R}^{d^l_{\text{in}}}$, and the columns form \emph{A tokens} $x^{A,l}_j = W^l_0[:,j] \in \mathbb{R}^{d^l_{\text{out}}}$. 
Since row and column tokens have different raw dimensionalities, we project them into a shared token space of width $d$:
\begin{equation}
\tilde{x}^{B,l}_i = x^{B,l}_i P_B^l,\quad P_B^l \in \mathbb{R}^{d_{\text{in}}^l \times d},\qquad
\tilde{x}^{A,l}_j = x^{A,l}_j P_A^l,\quad P_A^l \in \mathbb{R}^{d_{\text{out}}^l \times d}.
\end{equation}
We then concatenate them into a row/column token sequence:
\begin{equation}
X_W^l =
[\tilde{x}^{B,l}_1,\ldots,\tilde{x}^{B,l}_{d_{\text{out}}^l},
 \tilde{x}^{A,l}_1,\ldots,\tilde{x}^{A,l}_{d_{\text{in}}^l}]
\in \mathbb{R}^{(d_{\text{out}}^l+d_{\text{in}}^l)\times d}.
\end{equation}

\begin{wrapfigure}{r}{0.5\textwidth} % 'r' for right alignment, adjust 0.5 for width
    \centering
    \vspace{-2.5em} % Optional: Adjust vertical spacing if it pushes down too far
    \includegraphics[width=\linewidth]{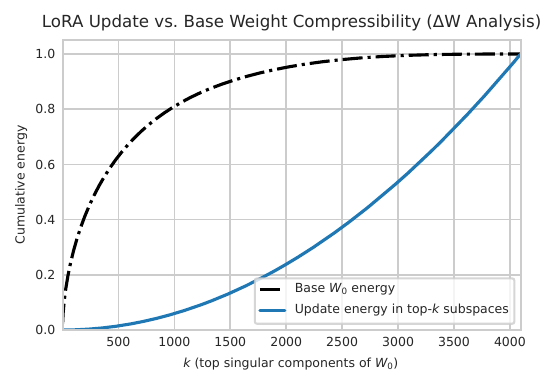}
    \vspace{-2em}
    \caption{
    Compressibility (measured by cumulative SVD energy) gap between the frozen base weights ($W_0$ in black) and LoRA updates ($\Delta W$ in blue). The plotted curves aggregate all LoRA layer pairs across the training set. While $W_0$ is highly compressible, $\Delta W$ spreads its energy across a much broader set of base singular directions, motivating full-rank base weight tokenization for accurate LoRA prediction.
    }
    \vspace{-1.5em}
    \label{fig:weight_analysis}
\end{wrapfigure}
This construction is inspired by classical CUR decompositions, which approximate a matrix using its actual columns and rows~\cite{hamm2019perspectivescurdecompositions,boutsidis2014optimalcurmatrixdecompositions}. We use the rows and columns of $W^l_0$ directly as tokens to expose the hypernetwork to the layer's native input/output feature coordinates and cross-feature correlations, instead of computing a CUR factorization here.

Since LoRA parameterizes adaptation as a low-rank update $\Delta W = BA$ defined with respect to the frozen base weight $W_0$, conditioning the predictor on $W_0$ provides a strong structural prior that is absent when mapping solely from text or noise embeddings. Empirically, we observe a clear compressibility gap: $W_0$ concentrates most of its spectral energy in the top singular components, while $\Delta W$ distributes its energy across a substantially broader set of singular directions (Fig.~\ref{fig:weight_analysis}).
This observation motivates the use of \emph{full-rank} weight tokenization of the base model as input to the hypernetwork. Aggressive compression of $W_0$ would discard singular directions that carry critical information for accurately predicting $\Delta W$. Refer to \cref{sec:ablation} for the ablation.

\noindent\textbf{Weight Token Compression.}
Although full-rank weight tokenization preserves critical structural directions, directly feeding the entire sequence $X^l_W$ into a hypernetwork is computationally prohibitive due to its large token length ($d^l_{\text{out}} + d^l_{\text{in}}$). 
Importantly, our goal is not to perform spectral truncation or low-rank approximation of $W_0$, which would discard singular directions that are informative for predicting $\Delta W$. Instead, we introduce a learnable token aggregation mechanism that reduces sequence length while still allowing the hypernetwork to attend to all original weight directions.
Specifically, we apply a cross-attention operation using learnable queries to aggregate the high-dimensional weight tokens:
\begin{equation}\label{eq:compress}
Q^l_W = \mathrm{CrossAttn}(Q_{\text{learned}},\, X^l_W,\, X^l_W),
\end{equation}
where $Q_{\text{learned}} \in \mathbb{R}^{n \times d}$ is a learnable query embedding, and the compressed weight token sequence $Q^l_W \in \mathbb{R}^{n \times d}$ has a much shorter token length $n \ll d^l_{\text{out}} + d^l_{\text{in}}$.

\noindent\textbf{SVD-Canonicalized LoRA Prediction.}
Instead of directly predicting the full unconstrained LoRA matrices $A^l$ and $B^l$ that are non-identifiable, we propose predicting the \textit{SVD-decomposed} LoRA weights. We pre-extract ground-truth LoRAs across diverse dynamic concepts and apply Singular Value Decomposition (SVD) to canonicalize the matrices $A^l$ and $B^l$:
\begin{equation}
\Delta W^l = U^lS^l(V^l)^T \Rightarrow B^{\star,l}=U^l(S^l)^{1/2}, A^{\star,l}=(S^l)^{1/2}(V^l)^T
\end{equation}
The hypernetwork is then trained to minimize the reconstruction loss against these SVD-canonicalized matrices. Because SVD enforces an orthogonal, energy-compacted basis, \textit{we empirically find that learning to predict these canonicalized matrices converges significantly faster and produces more stable generation than predicting raw, non-canonicalized weights}. See Sec.~\ref{sec:ablation} for the ablation.

\subsection{\methodname{} Architecture}
Designing a hypernetwork that generates high-dimensional adaptation weights 
for a large-scale video diffusion model is challenging due to the scale and 
layer-wise heterogeneity of the backbone. 
\methodname{} employs a Transformer-based hypernetwork $H_{\phi}$ that predicts 
all LoRA weights in a single forward pass, conditioned on the effect prompt 
and layer-specific metadata. $H_{\phi}$ consists of two parts: (i) a Transformer backbone $H_{\mathcal{E}}$ that fuses weightext and weight tokens, we inject semantics through cross-attention so that structurally-aware weight tokens attend to the effect semantics, and (ii) an attention-based weight prediction head $H_{\mathcal{D}}$ that uncompresses layer tokens and outputs LoRA factors.

\noindent\textbf{Transformer backbone $H_{\mathcal{E}}$.} 
$H_{\mathcal{E}}$ models the interaction between the structural priors of the base model and the semantic guidance of the effect prompt. For each adapted layer $l \in \{1,\dots,L\}$, the compressed weight tokens $Q_W^l$ serve as the primary input sequence. To provide spatial and structural awareness, we add learned layer-index embeddings $e_l$ and module-type embeddings $e^l_M$ directly to the base tokens:
\begin{equation}
Z^l = Q^l_W+e_l+e^l_M
\end{equation}
where $e_{l} = E_{\text{layer}}(l) \in \mathbb{R}^{d}$ and $e_{M}^{\,l} = E_{\text{mod}}\!\left(M^{l}\right) \in \mathbb{R}^{d}$.
The semantic effect prompt embedding $c_{\text{eff}}$ is encoded via a frozen text encoder 
to obtain a sequence of text tokens
$T_{\text{eff}} = E_{\text{text}}(c_{\text{eff}})\in\mathbb{R}^{n_t\times d}$.
Rather than concatenating text and weight tokens, we inject semantics through cross-attention so that structurally-aware weight tokens attend to the prompt.

$H_{\mathcal{E}}$ takes the layer-wise conditions as input and predicts the effect-injected latents used to generate LoRA weights:
\begin{equation}
X_{\mathcal{E}} = H_{\mathcal{E}}(Z, T_{\text{eff}}) 
\in \mathbb{R}^{L \times n \times d}.
\end{equation}
Here, $Z$ is the stacking of $\{Z^l\}^L_{l=1}$, $L$ denotes the number of adapted layers, $n$ is the number of latent tokens generated per layer, and $d$ is the hidden width of the hypernetwork. Each slice $X_{\mathcal{E}}^{\,l} \in \mathbb{R}^{n \times d}$ corresponds to the effect-injected latent associated with layer $l$, which is subsequently used to predict its LoRA weights.

\noindent\textbf{LoRA Weight Prediction Head $H_{\mathcal{D}}$.} The goal of $H_{\mathcal{D}}$ to predict \textit{full uncompressed weights of SVD-Canonicalized LoRA matrix} $A^\star$ and $B^\star$ from the compressed latent $X_\mathcal{E}$. $H_{\mathcal{D}}$ consists a cross-attention uncompression operation:
\begin{equation}
X^l_\mathcal{D} = \text{CrossAttn}(Q=X_W^l, K=X_\mathcal{E}^l, V=X_\mathcal{E}^l)\in\mathbb{R}^{(d^l_{\text{out}}+d^l_{\text{in}})\times d}
\end{equation}
This operation projects the learned effect dynamics directly back into the native structural dimensionality of the base weights by using the uncompressed original weights $X_W^l$ as queries.
We then split tokens into row-part and column-part:
\begin{equation}
X_{\mathcal{D}}^{\,l,(B)} = X_{\mathcal{D}}^{\,l}[\,:\,d_{\text{out}}^l], \qquad
X_{\mathcal{D}}^{\,l,(A)} = X_{\mathcal{D}}^{\,l}[\,d_{\text{out}}^l:\,].
\end{equation}
Finally, $H_{\mathcal{D}}$ employs two linear projection heads to predict the canonical LoRA matrices $\hat{B}$ and $\hat{A}$:
\begin{equation}
\hat{B}^{\,l}=\mathrm{Head}_B(X_{\mathcal{D}}^{\,l,(B)}) \in \mathbb{R}^{d_{\text{out}}^l \times r}, 
\quad 
\hat{A}^{\,l} = \left(\mathrm{Head}_A(X_{\mathcal{D}}^{\,l,(A)})\right)^{\top} \in \mathbb{R}^{r \times d_{\text{in}}^l},
\end{equation}

By processing all layer tokens jointly, $H_{\mathcal{D}}$ produces 
$\{(\hat{A}^l, \hat{B}^l)\}_{l=1}^{L}$ in a single forward pass, 
enabling efficient LoRA synthesis without per-layer optimization.

\noindent\textbf{Training Objective.} 
\label{sec:objective}
Given our predictions $\{(\hat{A}^{\,l}, \hat{B}^{\,l})\}_{l=1}^{L}$ and ground-truth SVD-canonicalized LoRA weights $\{(A^{\star,l}, B^{\star,l})\}_{l=1}^{L}$, we train $H_{\phi}$ using a \emph{normalized MSE (NMSE)} loss that measures the \emph{relative} Frobenius error for both $A$ and $B$:
\begin{equation}
\mathcal{L}_{\text{LoRA}}
=
\frac{1}{2L}
\sum_{l=1}^{L}
\left(
\frac{1}{N}\sum_{b=1}^{N}
\frac{\left\lVert \hat{A}^{\,l}_b - A^{\star,l}_b \right\rVert_F^2}
{\left\lVert A^{\star,l}_b \right\rVert_F^2 + \epsilon}
+
\frac{1}{N}\sum_{b=1}^{N}
\frac{\left\lVert \hat{B}^{\,l}_b - B^{\star,l}_b \right\rVert_F^2}
{\left\lVert B^{\star,l}_b \right\rVert_F^2 + \epsilon}
\right),
\end{equation}
where $b$ indexes the batch ($N$ samples) and $\epsilon$ is a small constant for numerical stability.
This relative formulation reduces sensitivity to scale variations across layers and effects, leading to more stable hypernetwork training.

At inference time, we directly apply the predicted LoRA parameters to the frozen video diffusion backbone to enable zero-shot effect generation without per-effect optimization. 
\section{Experiments}
\label{sec:experiments}

\subsection{Implementation Details}
\noindent\textbf{Hypernetwork.} Unless otherwise specified, we use a 1.3B-parameter Transformer hypernetwork (14 layers, hidden width $d=2048$) and set the target LoRA rank to $r=128$.. For each adapted layer, we compress weight tokens to $n=256$ latent tokens. It takes 3.3 seconds for our hypernetwork to generate the entire $\Delta W$ for the base model on A100. We primarily conduct experiments on our proprietary image-to-video (I2V) diffusion model. To validate that \methodname{} is model-agnostic, we additionally reproduce our core findings on the public WAN video model (see Appendix for details).

\noindent\textbf{Baselines.}
We compare against 
(i) the baseline I2V controlled using text prompt, 
(ii) a HyperDreamBooth-style hypernetwork baseline~\cite{ruiz2024hyperdreambooth} adapted to video LoRA prediction, and
(iii) a fully optimized LoRA trained per effect from scratch, which serves as an upper bound.
To ensure fairness, all hypernetworks predict LoRA updates for the same set of target layers and use the same training/evaluation data and inference settings.
% % Since the original HyperDreamBooth predicts all layers jointly and video backbones are substantially larger, we adapt its implementation to match our layer selection and prediction protocol.

\noindent\textbf{Data.} We compiled 75 effects prompts, each accompanied by approximately 50 manually selected, captioned videos. Roughly 40\% of this data was sourced from public video models, while the remainder was generated and collected internally. Five of the 75 prompts were reserved as a held-out evaluation set. Further details are provided in the Appendix.

\noindent\textbf{Training.} We train the hypernetwork using the AdamW optimizer across 4 nodes $\times$ 8 NVIDIA A100 GPUs for 4k epochs, with a constant learning rate of $5e^{-5}$ and a $0.01\%$ warmup ratio. For LoRA training, we utilize a single node for 1k steps, employing a cosine learning rate with an initial learning rate of $1e^{-4}$ with a $0.01$ warmup ratio and decaying to a minimum learning rate of $1e^{-5}$. 

\noindent\textbf{Evaluation.}
We use 70 in-distribution (IID) effects and 32 out-of-distribution (OOD) effects to evaluate our hypernetwork. For each effect, we generate 76 videos at $512\times288$ resolution using a fixed prompt suite spanning diverse subject categories (single/multi-person scenes, dynamic backgrounds, objects, and animals). We use identical prompts and random seeds across methods for generating the evaluation videos to ensure fairness.

We measure text-video alignment using CLIP Score~\cite{radford2021learning} computed between predicted and effect prompt, and overall video fidelity by Aesthetic Quality metric from VBench~\cite{huang2024vbench}. To directly assess whether the intended effect is executed (e.g., ``polar bear spa''), we employ Qwen3-VL~\cite{bai2025qwen3} as an expert VLM judge. The VLM scores three criteria: (i) \emph{Effect Execution}, (ii) \emph{Context Preservation}, and (iii) \emph{Temporal Quality}, and we report their sum as \emph{Overall}.

\subsection{Prompt2Effect Results}
We evaluate \methodname{} as a \emph{zero-shot} prompt-to-effect system that predicts LoRA weights in a single forward pass, and compare against the base model, HyperDreamBooth~\cite{ruiz2024hyperdreambooth}, and fully optimized LoRA (upper bound).

% \begin{wraptable}{r}{10cm}
\begin{table}[t]
\scriptsize
\centering
\caption{Quantitative results across evaluation metrics for IID effects. 
\textbf{Bold} denotes the best and \underline{underline} denotes the second best. 
The LoRA baseline that requires per-effect optimization is \textcolor{gray}{deemphasized}.}
\label{tab:quantitative_results}
% \resizebox{\linewidth}{!}{
\begin{tabular}{lccccccc}
\toprule
\multirow{2}{*}{Method} 
& \multirow{2}{*}{\makecell{CLIP \\ Score($\uparrow$)}} 
& \multirow{2}{*}{\makecell{Aesthetic \\ Quality($\uparrow$)}} 
& \multirow{2}{*}{\makecell{Motion \\ Smoothness($\uparrow$)}}
& \multicolumn{4}{c}{VLM Evaluation($\uparrow$)} \\
& & & &  Effect & Context & Temporal & Overall \\
\midrule

Baseline (Base I2V model) 
& 24.23 & 53.36 & 98.33 & 15.99 & 14.21 & 12.37 & 42.57 \\

HyperDreambooth~\cite{ruiz2024hyperdreambooth} 
& 22.34 & 35.30 & - & 0.00 & 0.00 & 0.00 & 0.00 \\

\textbf{\methodname{} (Ours) }
& \textbf{25.10} 
& \underline{56.30} 
& \textbf{98.54}
& \textbf{27.89} 
& \underline{15.34} 
& \underline{13.70} 
& \underline{56.93} \\
\rowcolor{gray!10} \color{gray}LoRA 
& \color{gray} \underline{25.06} 
& \color{gray} \textbf{57.60} 
& \color{gray} \underline{98.44}
& \color{gray} \underline{27.68} 
& \color{gray} \textbf{15.82} 
& \color{gray} \textbf{14.07} 
& \color{gray} \textbf{57.57} \\
\bottomrule
\end{tabular}
% }
% \end{wraptable}
\end{table}

\noindent\textbf{IID effects.}
Table~\ref{tab:quantitative_results} reports results on IID effects. \methodname{} substantially improves over the base model across all evaluation metrics and performs closely matching the LoRA upper bound. 
In contrast, the HyperDreamBooth~\cite{ruiz2024hyperdreambooth} baseline exhibits unstable training when scaled to video generation, leading to noisy outputs, near-zero VLM scores, and the lowest CLIP and aesthetic scores. These results highlight the effectiveness of our weight-driven, SVD-canonicalized hypernetwork design in enabling stable and high-quality weight prediction.

% \begin{wraptable}{r}{8cm}
\begin{table}[t]
\scriptsize
\centering
% \vspace{-1.5em}
\caption{Quantitative results of zero-shot and fast adaptation on OOD effects.}
\label{tab:ood}
% \resizebox{\linewidth}{!}{
\begin{tabular}{lccccccc}
\toprule
\multirow{2}{*}{Method} & \multirow{2}{*}{\makecell{CLIP \\ Score($\uparrow$)}} & \multirow{2}{*}{\makecell{Aesthetic \\ Quality($\uparrow$)}} & \multirow{2}{*}{\makecell{Motion \\ Smoothness($\uparrow$)}} & \multicolumn{4}{c}{VLM Evaluation($\uparrow$)} \\
                        & & & & Effect & Context & Temporal & Overall \\
\midrule
Baseline (Base I2V model)     & 23.17 & 52.26 & 98.42 & 15.24 & 14.78 & 12.44 & 42.46 \\
HyperDreambooth~\cite{ruiz2024hyperdreambooth} & 22.36 & 36.57 & - & 0.00     & 0.00     & 0.00     & 0.00     \\
\textbf{\methodname{} (Ours)}    & 23.52 & 53.44 & \textbf{98.58} & 20.50 & 15.42  & 12.95  & 48.87     \\
\midrule
Init-50         & 25.57 & 56.06 & 98.47 & 26.14 & 15.51 & 13.23 & 54.89 \\
Init-100        & \textbf{25.82} & \textbf{56.85} & \underline{98.53} & \underline{26.89} & \underline{15.65} & \underline{13.79} & \underline{56.33} \\
\rowcolor{gray!15}
\color{gray}
LoRA-100        & \color{gray} 25.11 & \color{gray} 56.50 & \color{gray} 98.20 & \color{gray} 24.66 & \color{gray} 14.70 & \color{gray} 12.91 & \color{gray} 52.28 \\
\rowcolor{gray!15}
\color{gray}
LoRA (1000 step) & \color{gray} \underline{25.73} & \color{gray} \underline{56.66} & \color{gray} 98.14 & \color{gray} \textbf{26.94} & \color{gray} \textbf{15.85} & \color{gray} \textbf{13.91} & \color{gray} \textbf{56.70} \\
\bottomrule
\end{tabular}
\vspace{-1.5em}
% \end{wraptable}
\end{table}

\begin{figure}[t]
    \centering
\includegraphics[width=\linewidth]{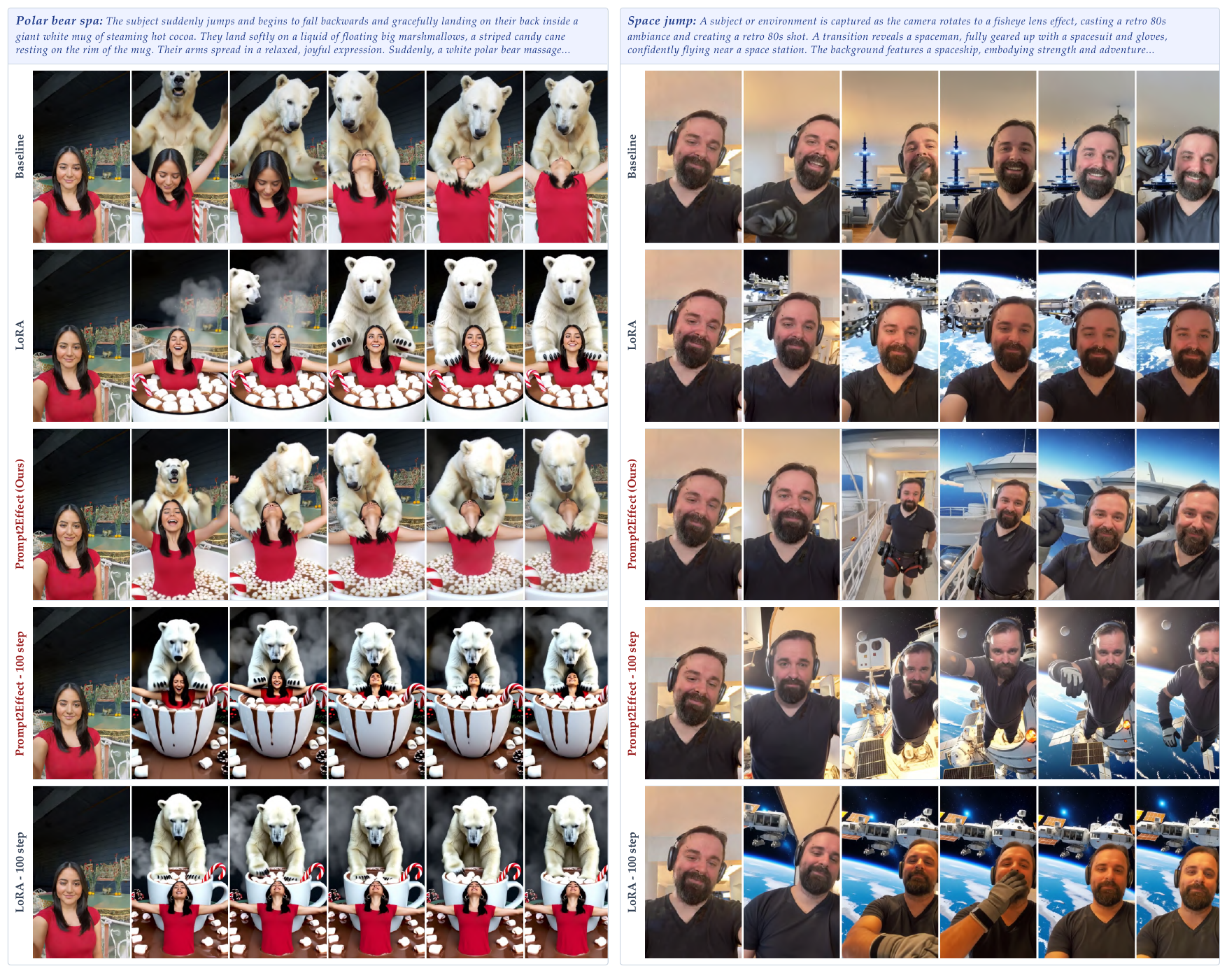}
\vspace{-2em}
\caption{Out-of-distribution (OOD) qualitative results and fast adaptation. We compare the base model, a fully optimized LoRA, our \textcolor{red!70!black}{\methodname{}}’s one-shot predicted weights (zero-shot), and 100-step adaptation initialized from our prediction (Init-100) versus 100-step LoRA training from scratch (LoRA-100). \methodname{}’s initialization substantially improves OOD effect within a small optimization budget, reducing $10\times$ training time compared to training LoRA from scratch (100 steps vs. 1000 steps).}
    \label{fig:main_ood}
\end{figure}

\noindent\textbf{OOD Effects.}  
While \methodname{} demonstrates strong zero-shot performance on IID concepts, highly out-of-distribution (OOD) effects may lack certain fine-grained, idiosyncratic details, although ours still outperforms both the base model and HyperDreamBooth, as shown in Table~\ref{tab:ood}. 
To further enhance OOD performance, we use the predicted weights as an initialization for subsequent LoRA optimization. Initializing from our zero-shot prediction (Init-50 and Init-100) significantly accelerates convergence compared to training from scratch (LoRA-100). Notably, with only 100 optimization steps (Init-100), our method approaches the performance of a fully converged standard LoRA model, which typically requires 1,000 steps. Specifically, Init-100 achieves an overall VLM score of 56.33, comparable to 56.70 for standard LoRA, while yielding a higher CLIP score (25.82 vs. 25.73). 
Qualitative results further show that this 100-step fast adaptation recovers fine-grained OOD textures that may be slightly smoothed in the zero-shot prediction, as illustrated in Fig.~\ref{fig:main_ood}. This indicates a $10\times$ training speed reduction using our \methodname{} as initialization compared to training LoRA from scratch.

\subsection{Public Model Experiments}
\begin{figure}[t]
    \centering
    \includegraphics[width=\linewidth]{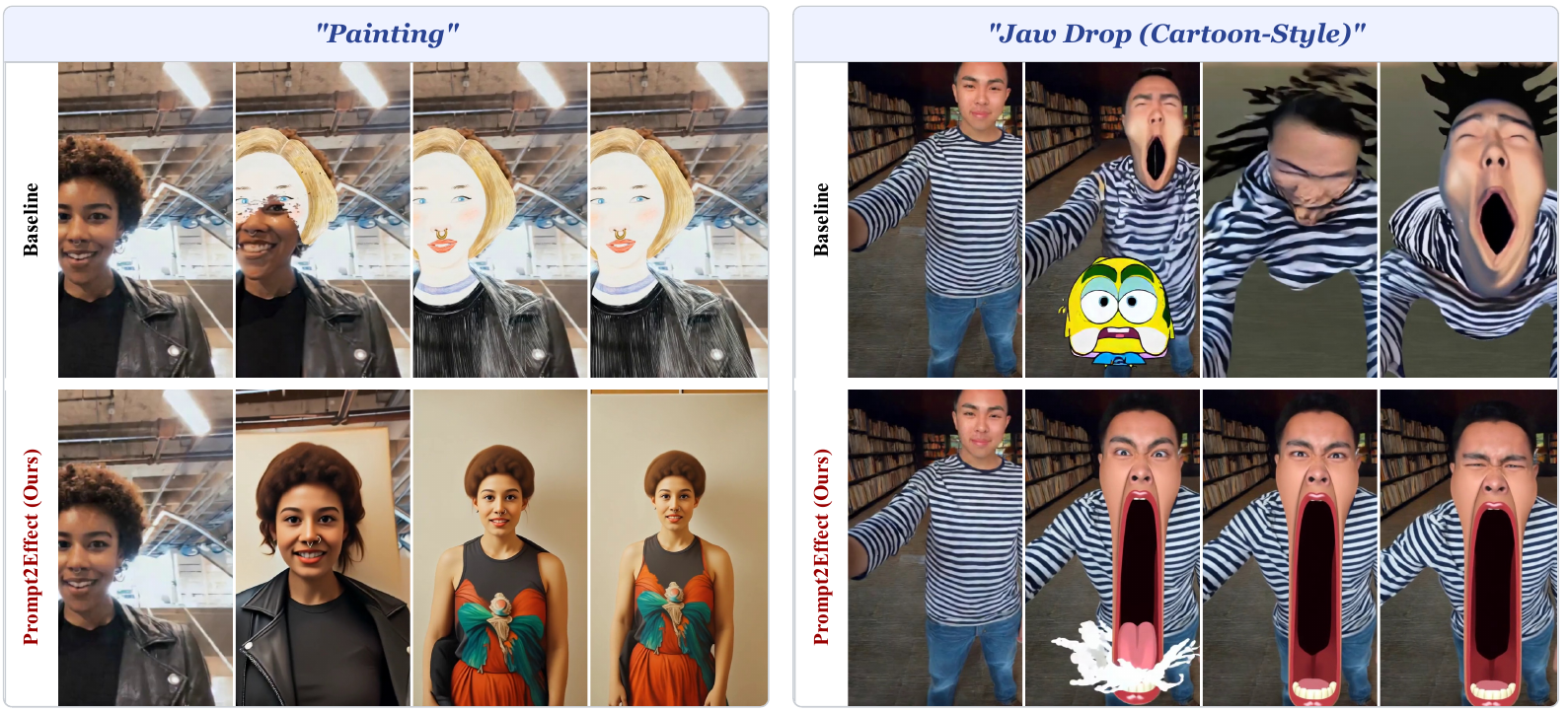}
    \caption{Qualitative effect generation applying \methodname{} on the Wan2.1-I2V-14B backbone.}
    \label{fig:wan_qualitative}
\end{figure}

To validate that the proposed Prompt2Effect is model-agnostic, we additionally reproduce our core findings on the public Wan2.1-I2V-14B-480P\footnote{\url{https://huggingface.co/Wan-AI/Wan2.1-I2V-14B-480P}}  model (see Appendix for details). As shown in \cref{tab:wan_results}, \methodname{} achieves highly competitive quantitative performance on the Wan backbone, closely tracking the fully optimized, per-effect LoRA baselines in both text-video alignment (CLIP Score) and effect execution (VLM Score), while actively improving motion smoothness over the baseline. Qualitative examples of these synthesized effects demonstrating the robust translation to the Wan architecture are provided in \cref{fig:wan_qualitative}.

\begin{table}[t]
\centering
\scriptsize
\caption{Quantitative results on the Wan2.1-I2V-14B-480P model.}
\label{tab:wan_results}
\begin{tabular}{lccccccc}
\toprule
\multirow{2}{*}{Method} 
& \multirow{2}{*}{\makecell{CLIP \\ Score($\uparrow$)}} 
& \multirow{2}{*}{\makecell{Aesthetic \\ Quality($\uparrow$)}} 
& \multirow{2}{*}{\makecell{Motion \\ Smoothness($\uparrow$)}} 
& \multicolumn{4}{c}{VLM Evaluation($\uparrow$)} \\
& & & & Effect & Context & Temporal & Overall \\
\midrule
Baseline (Wan2.1-I2V)
& 23.62 & 58.24 & 97.63 & 17.05 & 14.69 & 11.35 & 43.09 \\
\textbf{Prompt2Effect (Ours)}
& \textbf{24.79} & 62.31 & \textbf{98.02} & \textbf{34.51} & 15.90 & 14.08 & 64.49 \\
\rowcolor{gray!10} \color{gray}LoRA 
& \color{gray} 24.68 & \color{gray} \textbf{62.90} & \color{gray} 97.88 & \color{gray} 34.16 & \color{gray} \textbf{16.43} &\color{gray}  \textbf{14.57} & \color{gray} \textbf{65.16} \\
\bottomrule
\end{tabular}
\end{table}

\subsection{Ablation Study}\label{sec:ablation}

\noindent\textbf{HyperNetwork Design.} 
We investigate two architectural variants:
\begin{itemize}
\item \textit{Per-Block Network:} A modular design where separate, smaller hypernetworks (or heads) predict the weights for specific layers or blocks of the base model. This approach (denoted as ``Ensemble'' in Table~\ref{tab:ablation_design}) achieves suboptimal performance overall, suggesting that global coordination across layers is important for consistent effect synthesis.
\item \textit{Single Big Network:} Our unified, large-scale hypernetwork (1.3B parameters) models the joint distribution of all layer weights simultaneously. By routing features jointly across layers, this architecture achieves superior semantic consistency and higher generation quality.
\end{itemize}

\begin{wrapfigure}{r}{0.4\textwidth} % 'r' for right alignment, adjust 0.5 for width
    \centering
    \vspace{-2em} % Optional: Adjust vertical spacing if it pushes down too far
    \includegraphics[width=\linewidth]{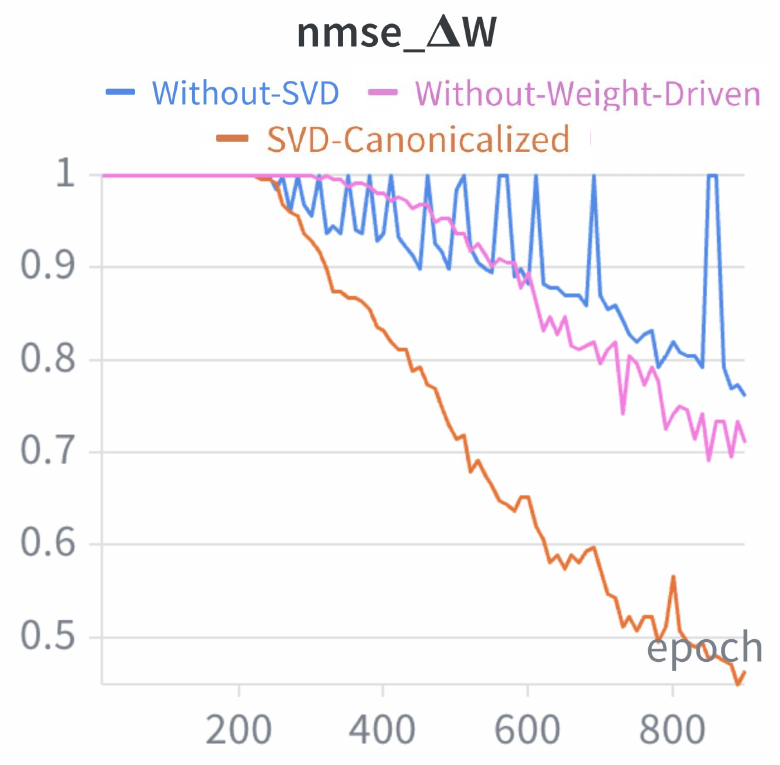}
    \vspace{-2em}
    \caption{Training curve to predict raw LoRA weights versus SVD-canonicalized prediction versus without weight-driven. 
    }
    \vspace{-1em}
    \label{fig:ablation_svd}
\end{wrapfigure}

\noindent\textbf{Weight-Driven Input Formulation.}
To verify the necessity of providing structural priors, we replace our weight-driven input formulation with standard abstract noise embeddings (denoted as ``Noise'' in Table~\ref{tab:ablation_design}). The noise-driven model struggles to map semantic text directly to functional weights, yielding a steep performance drop in VLM metrics (Overall: 46.14/43.58) and aesthetic quality. Training dynamics in \cref{fig:ablation_svd} further verify that, with the weight-driven input formulation, convergence is substantially accelerated. These confirm that exposing the hypernetwork to the principal subspaces of the base weights is crucial for predicting aligned low-rank updates. See Appendix for ablations of other designs other than weight-driven.

\noindent\textbf{SVD-Canonicalized Weight Prediction.}  
We compare training the hypernetwork to predict raw LoRA weights against predicting SVD-canonicalized targets $(A^\star, B^\star)$. As shown in Fig.~\ref{fig:ablation_svd}, SVD-canonicalization significantly accelerates convergence and improves optimization stability. The NMSE of the reconstructed update $\Delta W$ decreases smoothly throughout training, whereas the non-canonicalized variant exhibits pronounced oscillations and converges to a substantially higher final error.

\noindent\textbf{Compression Ratio.}
\cref{tab:ablation_compression} ablates the number of learned queries used to compress weight tokens ($n$).
We find that $n=256$ offers the best trade-off between compute and fidelity, outperforming $n=128$ and slightly improving over $n=512$ on average.

% \noindent\textbf{Layer Metadata Conditioning.}
% Removing the layer index $l$ and module type $M^l$ conditioning leads to catastrophic feature entanglement across different layers in our Single Big Network. Explicitly conditioning the hypernetwork on layer metadata is essential for enabling the model to accurately dispatch specific adaptation semantics to their corresponding network blocks.

\begin{table}[tb]
\small
\centering
\caption{Ablation study comparing Ensemble and Noise methods against our approach. Metrics are formatted as IID/OOD.}
\label{tab:ablation_design}
\resizebox{0.95\linewidth}{!}{
\begin{tabular}{lcccccc}
\toprule
\multirow{2}{*}{Method} & \multirow{2}{*}{\makecell{CLIP \\ Score}} & \multirow{2}{*}{\makecell{Aesthetic \\ Quality}} & \multicolumn{4}{c}{VLM Evaluation} \\
                        & & & Effect & Context & Temporal & Overall \\
\midrule
Ensemble & \underline{24.93}/\underline{23.46} & \underline{55.93}/\textbf{53.63} & \underline{23.21}/\underline{17.93} & \underline{14.9}/14.74 & \underline{13.19}/12.43 & \underline{51.29}/\underline{45.11} \\
Noise    & 24.41/23.21 & 53.83/52.21 & 19.35/15.99 & 14.33/\underline{14.92} & 12.45/\underline{12.67} & 46.14/43.58 \\
\textbf{Ours}     & \textbf{25.10}/\textbf{23.52} & \textbf{56.30}/\underline{53.44} & \textbf{27.89}/\textbf{20.50} & \textbf{15.34}/\textbf{15.42} & \textbf{13.70}/\textbf{12.95} & \textbf{56.93}/\textbf{48.87}  \\
\bottomrule
\end{tabular}
}
\end{table}

\begin{table}[tb]
\small
\centering
\caption{Ablation study comparing different compression ratio ($n$ in \cref{eq:compress}). Metrics are formatted as IID/OOD.}
\label{tab:ablation_compression}
\resizebox{0.95\linewidth}{!}{
\begin{tabular}{lcccccc}
\toprule
\multirow{2}{*}{Method} & \multirow{2}{*}{\makecell{CLIP \\ Score}} & \multirow{2}{*}{\makecell{Aesthetic \\ Quality}} & \multicolumn{4}{c}{VLM Evaluation} \\
                        & & & Effect & Context & Temporal & Overall \\
\midrule
n=128 & 24.23/23.17 & 53.21/52.76 & 21.10/17.54 & 15.29/14.84 & 13.36/12.63  & 49.75/45.01 \\
n=256 & \textbf{25.10}/\textbf{23.52} & \underline{56.30}/\underline{53.44} & \textbf{27.89}/\textbf{20.50} & \underline{15.34}/\underline{15.42} & \textbf{13.70}/\underline{12.95} & \textbf{56.93}/\textbf{48.87} \\
n=512 & \underline{24.83}/\underline{23.45} & \textbf{56.98}/\textbf{54.02} & \underline{27.25}/\underline{19.46} & \textbf{15.36}/\textbf{15.54} & \underline{13.61}/\textbf{13.05} & \underline{56.22}/\underline{48.03}\\
\bottomrule
\end{tabular}
}
\end{table}

\subsection{Zero-Shot Controllability}
\begin{figure}[t]
    \centering
\includegraphics[width=\linewidth]{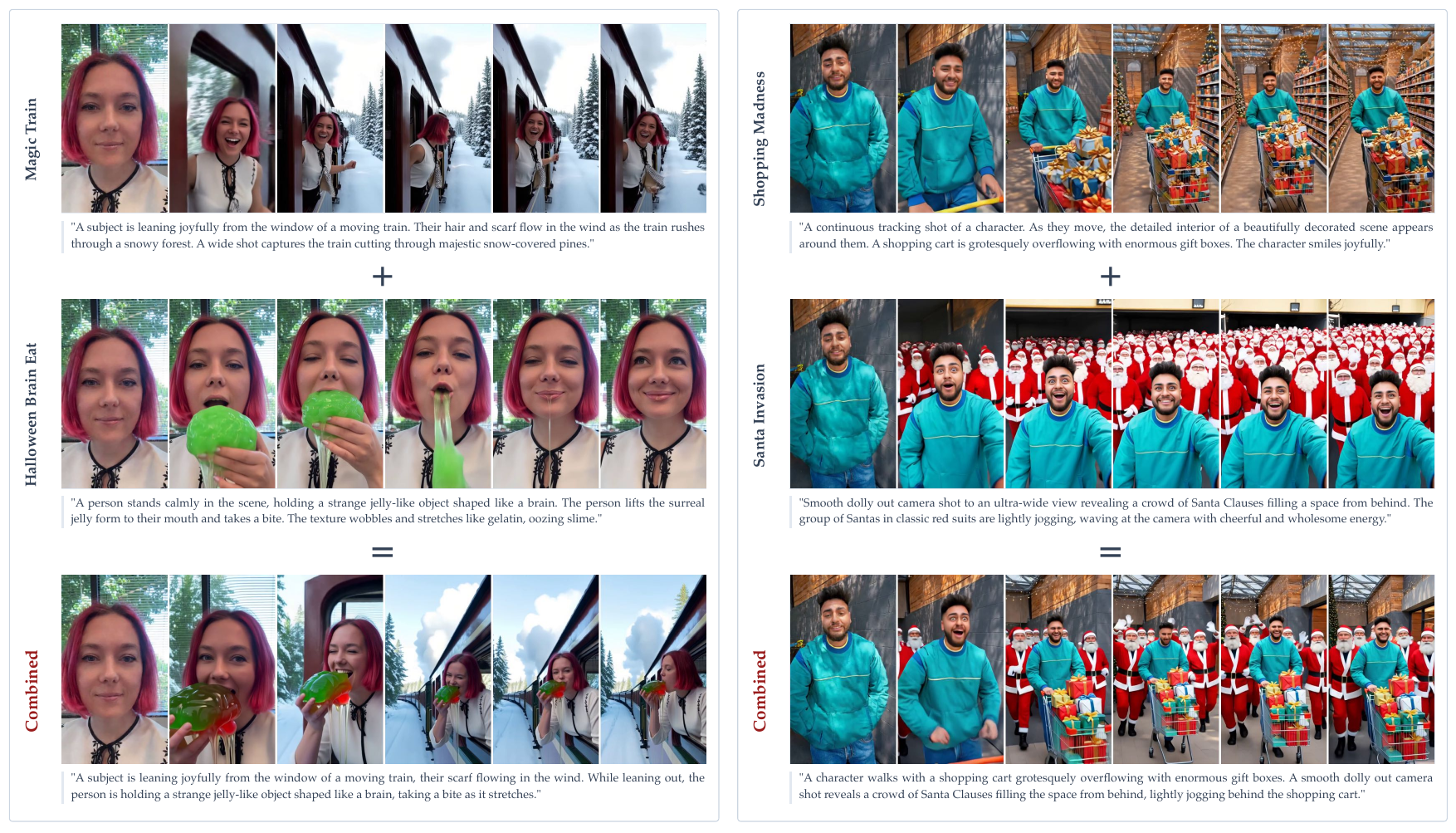}
\caption{Zero-shot compositional control by blending predicted LoRA updates, showing that \methodname{} learns a LoRA-like space supporting semantic composition.}
    \label{fig:composition}
\end{figure}

A key advantage of \methodname{} is that the synthesized adaptations behave like LoRA weights trained conventionally, enabling training-free zero-shot model control. We experiment with semantic composition by interpolating the predicted weights of two distinct effects ($\Delta W = 0.5 \Delta W_A + 0.5 \Delta W_B$). Visual results in \cref{fig:composition} confirm that our synthesized weights can be seamlessly blended, producing a combined video effect that exhibits characteristics of both concepts. 
% Furthermore, scaling the predicted weights by a constant multiplier smoothly controls the intensity of the generated effect, validating the robustness of our generated low-rank spaces.
\section{Conclusion}
In this work, we introduced \methodname{}, a training-free hypernetwork framework that synthesizes LoRA weights for controllable video generation in a single forward pass. By using a weight-driven input representation and predicting SVD-canonicalized adaptations, our method exploits the structural priors of large-scale video diffusion models. Extensive experiments show that \methodname{} achieves high-quality zero-shot video effects comparable to optimization-based LoRA, while reducing the computational cost from 56 GPU training hours to 3.3 seconds. Moreover, it provides a strong initialization that accelerates standard LoRA fine-tuning by up to $10\times$.
Despite these gains, out-of-distribution performance remains constrained by the diversity and scale of the training data for our hypernetwork. Future work will explore scaling data and training, and extending \methodname{} to stronger video backbones to further improve robustness and generalization.

\par\vfill\par
% % Now we have reached the maximum length of an ECCV \ECCVyear{} submission (excluding references and acknowledgements).
% % References should start immediately after the main text, but can continue past p.\ 14 if needed. 
% % \clearpage  % TODO FINAL: This \clearpage needs to be removed from both review and camera-ready versions.

% % \section*{Acknowledgements}
% % Please insert your acknowledgments here.

\bibliographystyle{splncs04}
\bibliography{main}

\clearpage
\appendix
\section{Public Model Experiments}
\label{sec:wan_experiments}

To validate that the proposed \methodname{} is fundamentally model-agnostic, we additionally reproduce our core findings on the open-weights Wan video model. 

\subsection{Experimental Setup}
\noindent\textbf{Base Model and Dataset:} We utilize the \texttt{Wan2.1-I2V-14B-480P} model
\footnote{\url{https://huggingface.co/Wan-AI/Wan2.1-I2V-14B-480P}} 
as our target image-to-video diffusion backbone. To construct the adaptation dataset, we collected 49 publicly available, high-quality community LoRAs with the same rank (32) curated from the Hugging Face \texttt{Remade-AI} collection
\footnote{\url{https://huggingface.co/collections/Remade-AI/wan21-14b-480p-i2v-loras}}.
This provides a diverse set of dynamic concepts to validate our approach on an entirely different 14B-parameter Diffusion Transformer architecture.

\noindent\textbf{Hypernetwork Configuration:} We maintain the exact same hypernetwork scale and architecture used for our proprietary model experiments: a 1.3B-parameter Transformer (14 layers, 32 attention heads, hidden dimension $d=2048$). The hypernetwork successfully compresses the massive base weights of the Wan 14B model into $n=256$ latent tokens per adapted layer. This proves that our architecture can efficiently scale to handle exceptionally large target models without requiring a proportional increase in the hypernetwork's parameter count.

\subsection{Experimental Results}
Consistent with our primary findings, applying our weight-driven input formulation and SVD-canonicalized prediction targets enabled stable convergence on the Wan 14B model. Despite the different base weight distributions and shapes from the Wan architecture, \methodname{} successfully synthesized LoRAs in a zero-shot manner. 

Furthermore, to comprehensively evaluate these models, we extend our metrics to include VBench Motion Smoothness at 480P. Because strict temporal consistency metrics can inadvertently penalize the intended dynamic transformations of a subject, appearance, or background, we report motion smoothness alongside our VLM-based temporal and effect scores. This combined approach directly measures whether the desired dynamic effect is executed coherently without unwarranted structural degradation.

\section{Dataset Details and Effect Prompts}
\label{sec:appendix_prompts}

\subsection{Data Collection and Prompts}
To train the hypernetwork effectively, we curated a comprehensive dataset of video effects consisting of 75 unique concepts (70 for training, 5 held-out for evaluation). These ground-truth LoRAs were initially trained on a state-of-the-art proprietary image-to-video diffusion model.
We further gather 27 validation effects from independent sources (public effects, synthesized, etc.) to make up a 32-effect OOD testing set. 

Our automated data collection pipeline leverages large language models (LLMs) and vision-language models (VLMs) to generate paired training data:
\begin{enumerate}
\item An LLM generates detailed effect/LoRA descriptions alongside corresponding first-frame image prompts.
\item A Text-to-Image (T2I) model, such as FLUX, renders the high-quality first frame.
\item An LLM/VLM takes the first frame and effect description to generate Image-to-Video (I2V) or First-Frame-to-Video (FLF2V) prompts.
\item Public video generation models (e.g., Wan) and our proprietary model renders the target video.
\item We apply aggressive filtering by human annotators to only keep the data with the highest quality and prompt alignment. 
\end{enumerate}
% This workflow provides the diverse dataset needed for large-scale dynamic concept pretraining.

The 75 curated effects span a wide variety of dynamic concepts, which we broadly categorize into: \textbf{Transformations} (e.g., \textit{skull\_reveal}, \textit{holiday\_elf}), \textbf{Camera \& Stylization} (e.g., \textit{fisheye\_animation}, \textit{dream\_cute\_sketch}), \textbf{Surreal Actions} (e.g., \textit{polarbear\_ride}, \textit{magic\_train}), and \textbf{Atmospheric Moods} (e.g., \textit{dark\_mood}, \textit{yellow\_mood}). Each effect prompt is accompanied by approximately 50 manually selected, captioned videos. Approximately 40\% of these videos were sourced from public video models, while the remaining 60\% were generated and collected internally. For the complete list of all 75 effect prompts and their full text descriptions, please refer to the list below.

% Our dataset consists of 75 unique dynamic effect concepts. We used 70 of these for training our Prompt2Effect hypernetwork and held out 5 for out-of-distribution (OOD) zero-shot evaluation. 

\subsection{Base Weight Compressibility Analysis}
\label{sec:fig3_details}
In Figure 3 of the main manuscript, we illustrated the compressibility gap between the frozen base weights ($W_0$) and the LoRA updates ($\Delta W$). To ensure a robust and representative analysis, the plotted curves do not represent a single layer or a single effect. Instead, they aggregate \emph{all} valid LoRA--layer pairs across our entire training set of 70 dynamic effects.

Specifically, for each adapted layer across all trained effect LoRAs, we performed Singular Value Decomposition (SVD) on the base weight matrix $W_0$ to extract its top-$k$ singular components. We then computed the cumulative energy of the LoRA update $\Delta W$ when projected into these top-$k$ subspaces. The solid curves presented in Figure 3 depict the mean cumulative energy computed across all aggregated LoRA layer combinations, while the shaded bounding regions represent the standard deviation. This comprehensive aggregation confirms that the observed compressibility gap, where $\Delta W$ distributes its energy much more broadly than $W_0$, is a universal property across our effect library, motivating our full-rank base weight tokenization strategy.

\subsection{List of Effect Concepts}
Below is the complete list of the 75 effect concepts curated for this study. 

\begin{tcolorbox}[
    breakable,
    colback=gray!5,       % Light gray background so it stands out like a figure
    colframe=black!75,    % Dark gray border
    title={\textbf{Training Effects (70)}},
    boxrule=0.5pt,
    arc=2pt,              % Slightly rounded corners
    left=4pt, right=4pt, top=4pt, bottom=4pt
]
\footnotesize % Shrink the font slightly to fit the long underscore names
\begin{multicols}{2} % Use 2 columns instead of 3 to prevent overfull rows
\begin{itemize}\setlength\itemsep{0em}
    \item body\_dance
    \item cat\_spa
    \item hair\_growth
    \item skull\_reveal
    \item head\_spin
    \item shaving
    \item morning\_coffee
    \item babyface
    \item balloon
    \item siblings
    \item trap
    \item twist
    \item transformation
    \item hit
    \item bug\_eat
    \item levitation
    \item chickenride
    \item halloween\_creatures
    \item magic\_train
    \item young\_and\_old
    \item futuristic\_car
    \item thanksgiving\_chef
    \item santa\_helper
    \item shoppingmadness
    \item santa\_crime
    \item magic\_ride
    \item rope\_walk
    \item robot\_under\_skin
    \item hw\_clown\_transform
    \item broomv3
    \item halloween\_brain\_eat
    \item utility\_wind
    \item cowride
    \item halloween\_green
    \item soft\_cartoon
    \item sled\_ride
    \item stolen\_by\_santa
    \item santa\_invasion
    \item make\_up
    \item holiday\_skating
    \item polarbear\_ride
    \item panda
    \item cartoon\_transform
    \item fov\_change\_seagull
    \item christmas\_cozy
    \item holiday\_elf
    \item holiday\_cookie
    \item yeti
    \item 3d\_tween\_dream
    \item timelapse
    \item retro\_game\_animation
    \item retr\_gam
    \item yellow\_mood
    \item red\_mood
    \item pink\_mood
    \item purple\_mood
    \item dark\_mood
    \item dream\_cute\_sketch\_v1
    \item dream\_cute\_sketch\_v2
    \item dream\_cute\_sketch\_v3
    \item cartoon\_style\_1
    \item cartoon\_style\_2
    \item cartoon\_style\_3
    \item cartoon\_style\_4
    \item cartoon\_style\_5
    \item nature\_inside
    \item clay\_animation
    \item clay\_transformation
    \item elegant\_timelapse
    \item fisheye\_animation
\end{itemize}
\end{multicols}
\end{tcolorbox}

\begin{tcolorbox}[
    breakable,
    colback=gray!5,       % Light gray background
    colframe=black!75,    % Dark gray border
    title={\textbf{Held-Out Evaluation Effects (5)}},
    boxrule=0.5pt,
    arc=2pt,              % Slightly rounded corners
    left=4pt, right=4pt, top=4pt, bottom=4pt
]
\footnotesize
\begin{itemize}\setlength\itemsep{0em}
    \item polarbear\_spa
    \item dj\_cat
    \item blue\_mood
    \item dream\_cute\_sketch\_v4
    \item fisheye\_transformation
\end{itemize}
\end{tcolorbox}

\subsection{Example Full-Text Prompts}
To illustrate the level of detail and structural guidance provided to the generation models, Table~\ref{tab:example_prompts} presents the complete text descriptions for a representative subset of our effects.

\begin{table}[h!]
\centering
\caption{Representative examples of the detailed text prompts used to define dynamic concepts in our dataset.}
\label{tab:example_prompts}
\renewcommand{\arraystretch}{1.3} 
\begin{tabular}{@{}p{0.26\linewidth}p{0.70\linewidth}@{}}
\toprule
\textbf{Effect Name} & \textbf{Full Prompt Description} \\ \midrule
\texttt{dj\_cat} & A main subject stands centered in the frame or is in the scene. The camera slowly and smoothly zooms out, gradually revealing more of the environment. As the frame widens, it clearly reveals that a DJ deck and a realistic-looking cat wearing headphones is already positioned nearby... The cat is actively DJ’ing from the start, confidently moving its paws on the controls as if mixing music. \\ \midrule
\texttt{polarbear\_spa} & The subject suddenly jumps and begins to fall backwards and gracefully land on their back inside a giant white mug of steaming hot cocoa. They land softly on a liquid of floating big marshmallows... Suddenly, a white polar bear with realistic fur comes up from the side behind the person and gives a gentle shoulder massage. \\ \midrule
\texttt{skull\_reveal} & A subject, group of people, object, or animal stands in the scene, or a background sets the scene. Suddenly, the skin begins to melt down from the skull, or the object begins to melt down, or the atmosphere shifts. This reveals a hyper-realistic skull or a hyper-realistic animal skull underneath... \\ \midrule
\texttt{fisheye\_animation} & The subject transforms into an extreme fisheye lens look with strong wide-angle distortion. They smile confidently and stylishly, often showing teeth where luxury diamond braces or reflections catch the light... Cinematic lighting, high contrast, and glossy reflections create an aggressive yet elegant hip-hop vibe. \\
\bottomrule
\end{tabular}
\end{table}

\section{Detailed Evaluation Metrics}
\label{sec:appendix_metrics}

To provide a comprehensive assessment of our training-free LoRA synthesis, we evaluate the generated videos across semantic alignment, visual fidelity, and complex effect execution. Below, we detail the specific protocols and formulations for each metric.

\subsection{Automated Metrics: CLIP Score and Aesthetic Quality}

\noindent\textbf{CLIP Score:} Standard text-to-image alignment metrics must be adapted to account for the temporal dimension of video. We measure text-video alignment using the \texttt{ViT-B/16} variant of the CLIP model~\cite{radford2021learning}. To ensure the semantic concept is maintained throughout the motion, we uniformly sample 16 frames across the duration of each generated video. We compute the cosine similarity between the CLIP text embedding of the intended effect prompt and the visual embedding of each individual frame. These frame-level similarities are scaled by a factor of 100 and averaged to produce the final video-level score. This multi-frame averaging ensures that models are penalized if an effect appears only momentarily or degrades over time.

\noindent\textbf{Aesthetic Quality:} While CLIP Score measures context alignment, it does not account for visual artifacts or degradation. To evaluate the overall perceptual fidelity, we report the Aesthetic Quality metric from the VBench~\cite{huang2024vbench} suite. This ensures that the injection of highly dynamic effects via Prompt2Effect does not compromise the underlying generative quality and structural integrity of the base video diffusion model.

\subsection{VLM Evaluation Protocol}

Automated metrics like CLIP and VBench often struggle to evaluate complex, sequential events (e.g., verifying if a "slow transformation" happens gradually rather than as an instantaneous jump-cut). To directly assess the success of the applied effects, we employ Qwen3-VL~\cite{bai2025qwen3} as an expert Vision-Language Model (VLM) judge. 

The VLM evaluates each video on a 100-point scale distributed across three criteria: Effect Execution (60 points), Identity \& Context Preservation (20 points), and Temporal \& Physics Quality (20 points). To ensure highly standardized, unbiased, and reproducible scoring, we query the VLM using a strict system prompt. The prompt includes explicit "Hard Fail" caps to heavily penalize severe visual corruption or complete failure of the core LoRA action. 

The complete system prompt provided to the VLM is detailed in Prompt~\ref{prompt:vlm_judge}. The model's responses were constrained to output valid JSON for automated parsing.

\vspace{1em} % Adds a little breathing room before the box

\begin{tcolorbox}[
    breakable, % THIS IS THE MAGIC WORD
    colback=gray!5,
    colframe=black!75,
    title={Prompt 1: Qwen3-VL Evaluation Template},
    label=prompt:vlm_judge
]
\small
\textbf{System Prompt:}\\
You are an expert judge evaluating AI-generated videos powered by specific "LoRA" action and transformation effects.

\textbf{INPUTS:}
\begin{itemize}
    \item \textbf{LORA EFFECT NAME:} "\{effect\_name\}"
    \item \textbf{LORA EFFECT PROMPT:} "\{target\_prompt\}"
    \item \textbf{TEST VIDEO:} The provided video to evaluate.
\end{itemize}

\textbf{EVALUATION RUBRIC (Total 100 points):}
\begin{enumerate}
    \item \textbf{EFFECT EXECUTION \& SEQUENCE (0-60 points):} Does the video perform the exact sequence of events described in the prompt? If the prompt asks for a "slow, smooth transformation", the video MUST demonstrate a gradual morphing. Harsh jump-cuts should be heavily penalized here.
    \item \textbf{IDENTITY \& CONTEXT PRESERVATION (0-20 points):} Does the main subject maintain their base identity (face, clothing, body) throughout the action, except where commanded? Does the background remain stable without unintended morphing or warping?
    \item \textbf{TEMPORAL \& PHYSICS QUALITY (0-20 points):} Is the motion fluid and natural? Are there minimal AI artifacts or flickering?
\end{enumerate}

\textbf{HARD FAIL CAPS (Force FINAL SCORE $\leq$ 20 if ANY are true):}
\begin{itemize}
    \item The core LoRA action completely fails to happen.
    \item The subject's intended base identity is completely destroyed.
    \item Severe visual corruption.
\end{itemize}

\textbf{EXAMPLES:}

\textit{[BAD CASE EXAMPLE VIDEO]}\\
\textbf{LoRA Prompt:} "A character stands in a high-fashion studio setup. The head of the character begins a slow, smooth transformation into an anthropomorphic animal head... while the rest of the scene remains unchanged."\\
\textbf{Reasoning:} The effect execution is poor (15/60) because the head changes instantly via a jump-cut, missing the "slow, smooth" sequence. Identity preservation is failing (4/20) because the clothing also morphs. Temporal quality is low (8/20). Triggers Hard Fail.\\
\textbf{JSON:} \texttt{\{"reasoning": "Head transforms instantly instead of slowly. Clothing morphs unintentionally.", "effect\_execution": 15, "identity\_preservation": 4, "temporal\_quality": 8, "hard\_fail": true\}}

\vspace{0.5em}
\textit{[GOOD CASE EXAMPLE VIDEO]}\\
\textbf{LoRA Prompt:} "A character stands in a high-fashion studio setup. The head of the character begins a slow, smooth transformation into an anthropomorphic animal head... while the rest of the scene remains unchanged."\\
\textbf{Reasoning:} The effect execution is excellent (58/60) showcasing a gradual, elegant transition. Identity preservation is perfect (19/20) as the clothing and background remain stable. Temporal physics are smooth (18/20).\\
\textbf{JSON:} \texttt{\{"reasoning": "Beautiful, slow head transformation. High-fashion suit and studio background remain perfectly stable.", "effect\_execution": 58, "identity\_preservation": 19, "temporal\_quality": 18, "hard\_fail": false\}}
\end{tcolorbox}

\section{Implementation Details}
\label{sec:implementation}

We summarize our architectural and training configurations in Table~\ref{tab:implementation_details} and provide detailed descriptions below.

\noindent\textbf{Hypernetwork Architecture:} We parameterize Prompt2Effect using a 1.3B-parameter Transformer hypernetwork consisting of 14 layers, 32 attention heads, and a hidden dimension of $d=2048$. To handle the massive dimensionality of video diffusion weights, we compress the input base weight tokens into $n=256$ latent tokens using a cross-attention compressor block. Prior to compression, the base weight row and column tokens are augmented with structural conditioning through 1D sinusoidal positional encodings and learned layer and module-type embeddings. The core backbone processes these $n$ latents through 14 sequential blocks, each containing a multi-head self-attention layer followed by a cross-attention layer that injects the text prompt embeddings. Finally, a cross-attention decompressor uses the original uncompressed weight tokens as queries to uncompress the latents, which are then mapped to the target LoRA rank via two linear projection heads. At 1.3B parameters, Prompt2Effect represents approximately $12\%$ of the parameters of the 11B base video diffusion model it adapts, making it a highly scalable solution for large-scale architectures.

\noindent\textbf{Optimization:} The hypernetwork is trained using the AdamW optimizer (with a default weight decay of $0.01$) across 32 NVIDIA A100 GPUs (4 nodes $\times$ 8 GPUs). We train for 4,000 epochs with a per-GPU batch size of 20 (resulting in a total global batch size of 640). We employ a constant learning rate schedule of $5 \times 10^{-5}$ following a $0.01\%$ linear warmup period. To ensure training stability across the large parameter space, we utilize \texttt{bfloat16} mixed precision and apply gradient clipping with a maximum L2 norm of 1.0. The model is optimized using a Normalized Mean Squared Error (NMSE) loss computed between the predicted and ground-truth SVD-canonicalized LoRA factors.

\noindent\textbf{Inference:} At inference time, it takes approximately 3.3 seconds for our hypernetwork to generate the entire set of $\Delta W$ updates for the base model on a single A100 GPU.

\begin{table}[h]
\centering
\caption{Detailed architectural and optimization hyperparameters for Prompt2Effect.}
\label{tab:implementation_details}
\small
\begin{tabular}{@{}ll@{}}
\toprule
\textbf{Hyperparameter / Setting} & \textbf{Value} \\ \midrule
\multicolumn{2}{@{}l}{\textit{Architecture}} \\ \midrule
Total Parameters & 1.3B \\
Transformer Layers & 14 \\
Attention Heads & 32 \\
Hidden Dimension ($d$) & 2048 \\
Latent Tokens ($n$) & 256 \\
Structural Conditioning & Sinusoidal PE, Layer \& Module embeddings \\ \midrule
\multicolumn{2}{@{}l}{\textit{Training \& Optimization}} \\ \midrule
Hardware & 32$\times$ NVIDIA A100 GPUs (4 nodes) \\
Optimizer & AdamW (Weight Decay $= 0.01$) \\
Training Epochs & 4,000 \\
Batch Size & 20 per GPU (Global Batch Size: 640) \\
Learning Rate & $5 \times 10^{-5}$ \\
Learning Rate Schedule & Constant with $0.01\%$ Linear Warmup \\
Mixed Precision & \texttt{bfloat16} \\
Gradient Clipping & 1.0 (L2 Norm) \\
Loss Function & NMSE on SVD-canonicalized factors \\ \midrule
\multicolumn{2}{@{}l}{\textit{Inference}} \\ \midrule
Hardware & 1$\times$ NVIDIA A100 GPU \\
Latency & $\sim$3.3 seconds per effect \\ \bottomrule
\end{tabular}
\end{table} 

\section{Additional Results and Ablations}
\label{sec:additional_results}

\subsection{Additional Ablations}
\noindent\textbf{Other Input Designs:} 
As referenced in the main text, we compare our weight-driven input formulation with standard abstract noise embeddings. The noise-driven model struggles to map semantic text directly to functional weights, yielding a steep performance drop.

\begin{wrapfigure}{r}{0.5\textwidth} % 'r' for right alignment, adjust 0.5 for width
    \centering
    \vspace{-2.5em} 
    \includegraphics[width=\linewidth]{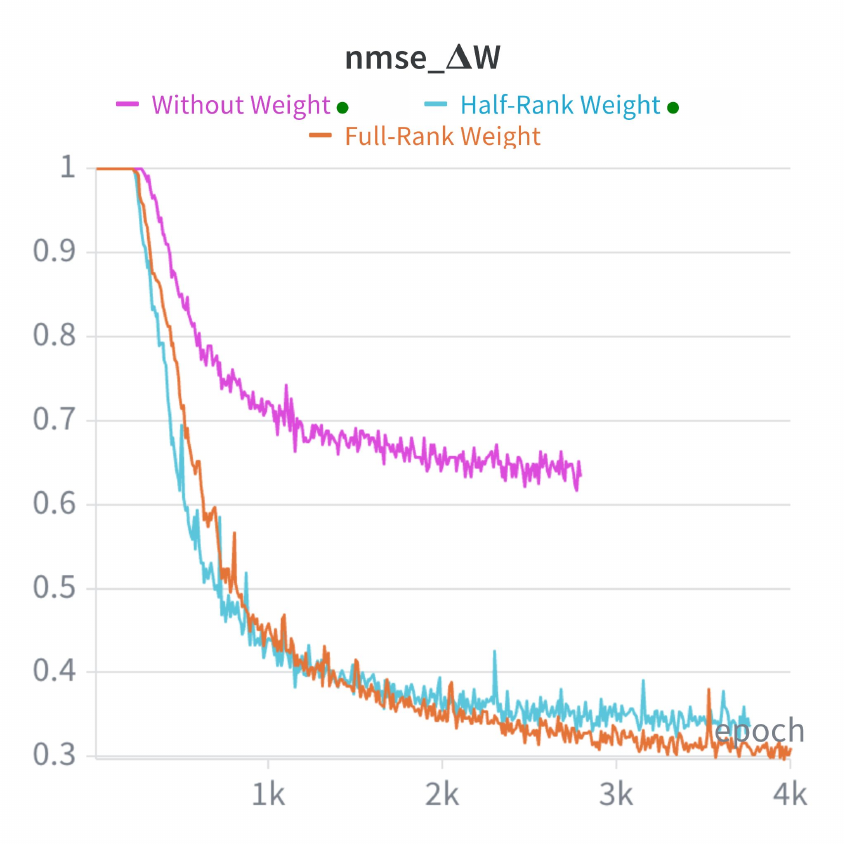}
    \vspace{-2em} 
    \caption{Training convergence of input designs. Our ``Full-Rank Weight'' formulation (orange) achieves the most stable convergence and lowest final NMSE compared to the ``Without Weight'' (pink) and ``Half-Rank Weight'' (blue) baselines.
    }
    \label{fig:appendix_ablation}
    \vspace{-2em}
\end{wrapfigure}

To further demonstrate the necessity of providing the hypernetwork with the complete structural priors of the base model, we ablate two additional input architectures (with training convergence shown in Figure~\ref{fig:appendix_ablation}):
\begin{itemize}
    \item \textbf{Without Weight:} We completely remove the \texttt{base\_weight} input, replacing the weight encoders with learned static structural queries and bypassing the compression module. This model maps semantic text directly to weight updates. Without the structural grounding of the base weights, this model struggles severely, plateauing at a high reconstruction error (NMSE $\approx 0.65$, pink curve in Figure~\ref{fig:appendix_ablation}).
    \item \textbf{Truncated SVD (Half-Rank) Weight:} Instead of passing the full-rank base weight $W_0$, we pass truncated SVD factors ($U_r$, $V_r$) to the hypernetwork's encoders. While this preserves the principal components of the layer and converges significantly better than the weightless model (NMSE $\approx 0.35$, blue curve in Figure~\ref{fig:appendix_ablation}), it discards lower-energy singular directions. These discarded directions still carry critical structural information required for accurately predicting the target LoRA updates.
\end{itemize}

As summarized in Table~\ref{tab:ablation_design_appendix} and Figure~\ref{fig:appendix_ablation}, our \textbf{Full-Rank Weight} approach, which ingests the uncompressed $W_0$ and uses cross-attention to dynamically compress it, achieves the most stable convergence and the lowest final error (NMSE $\approx 0.30$).

\begin{table}[h!]
\centering
\caption{Ablation of hypernetwork input designs. Utilizing the full-rank base weights as structural conditioning yields the lowest Normalized Mean Squared Error (NMSE) during training.}
\label{tab:ablation_design_appendix}
\small
\renewcommand{\arraystretch}{1.2}
\begin{tabular}{@{}llc@{}}
\toprule
\textbf{Model Design} & \textbf{Input Representation} & \textbf{Final NMSE ($\downarrow$)} \\ \midrule
Without Weight & Learned Static Queries & $\sim 0.65$ \\
Half-Rank Weight & Truncated SVD Factors ($U_r, V_r$) & $\sim 0.35$ \\
Full-Rank Weight (Ours) & Full Base Weight Matrix ($W_0$) & $\mathbf{\sim 0.30}$ \\ \bottomrule
\end{tabular}
\end{table}

% \begin{table}[tb]
% \small
% \centering
% \caption{Ablation study on additional design choices (e.g., text-driven vs. weight-driven). Metrics are formatted as IID/OOD.}
% \label{tab:ablation_design_appendix}
% \resizebox{0.95\linewidth}{!}{
% \begin{tabular}{lcccccc}
% \toprule
% \multirow{2}{*}{Method} & \multirow{2}{*}{\makecell{CLIP \\ Score}} & \multirow{2}{*}{\makecell{Aesthetic \\ Quality}} & \multicolumn{4}{c}{VLM Evaluation} \\
%                         & & & Effect & Context & Temporal & Overall \\
% \midrule
% [PLACEHOLDER: Baseline 1] & - & - & - & - & - & - \\
% [PLACEHOLDER: Baseline 2] & - & - & - & - & - & - \\
% \textbf{Ours}     & \textbf{25.10}/\textbf{23.52} & \textbf{56.30}/\underline{53.44} & \textbf{27.89}/\textbf{20.50} & \textbf{15.34}/\textbf{15.42} & \textbf{13.70}/\textbf{12.95} & \textbf{56.93}/\textbf{48.87}  \\
% \bottomrule
% \end{tabular}
% }
% \end{table}

\subsection{Additional Qualitative Results}
\begin{figure}[t]
    \centering
\includegraphics[width=0.7\linewidth]{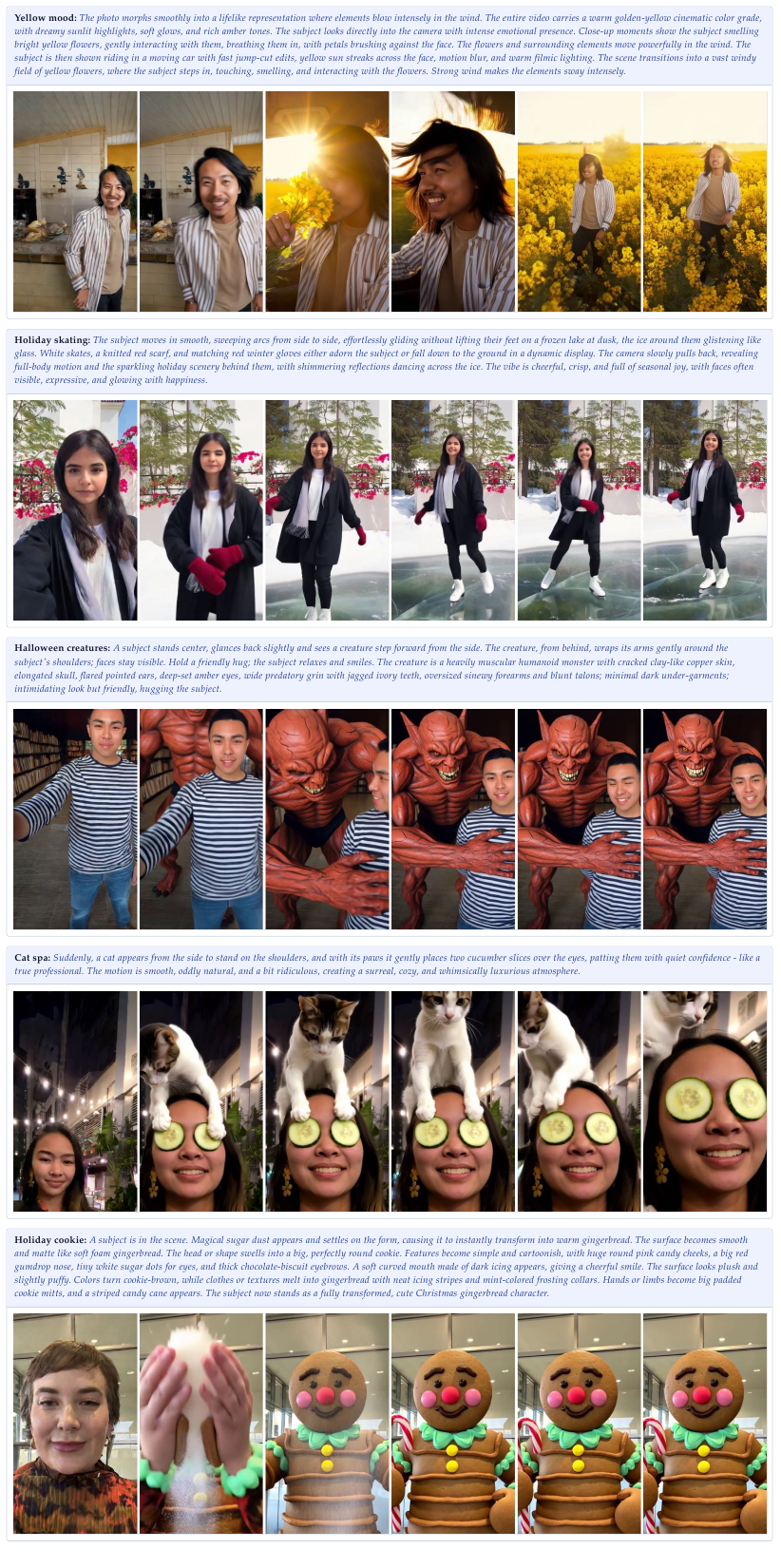}
\vspace{-2em}
\caption{Prompt2Effect's in-distribution effects prediction visualization}
    \label{fig:iid_viz1}
\end{figure}

\begin{figure}[t]
    \centering
\includegraphics[width=0.9\linewidth]{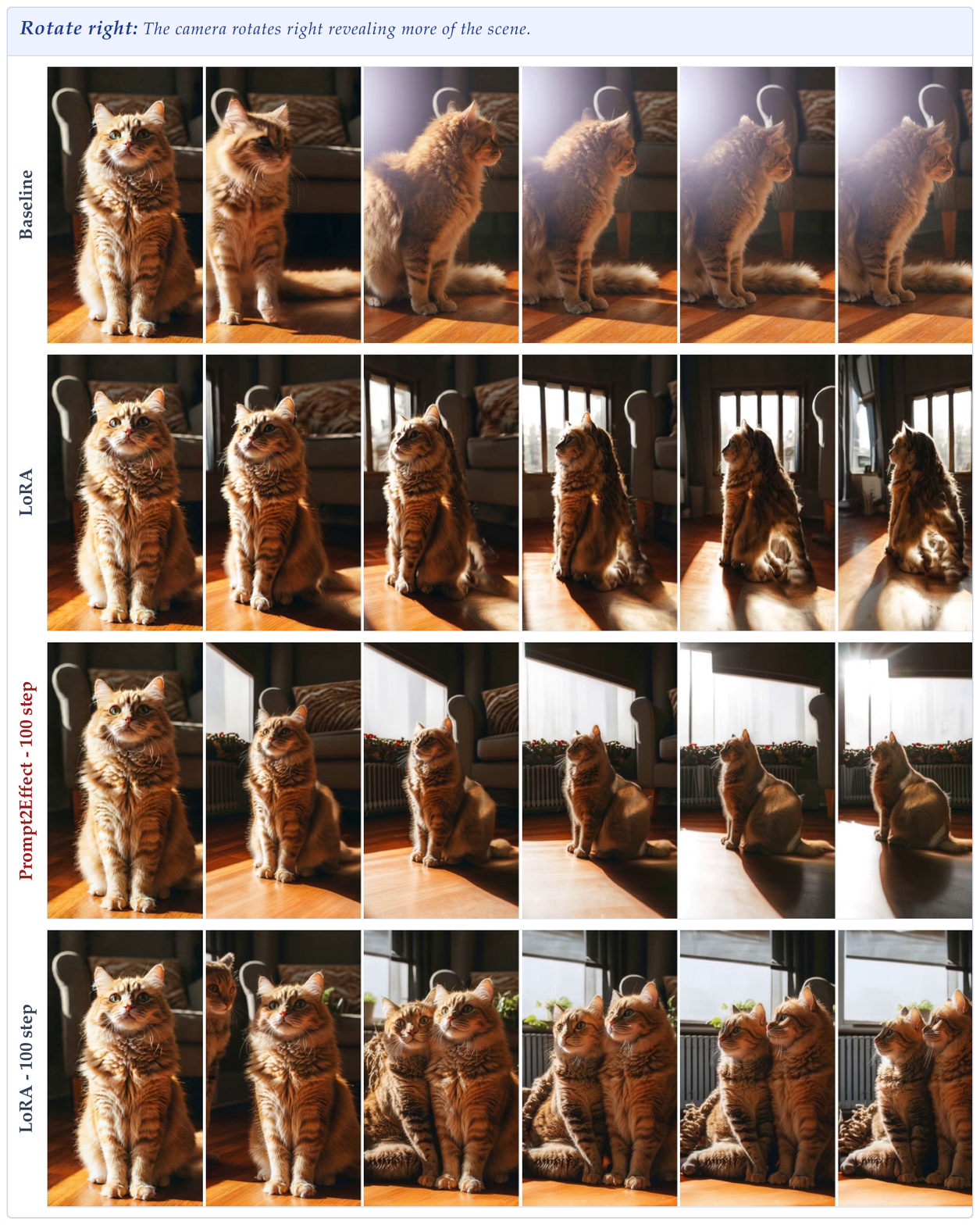}
\vspace{-2em}
\caption{More visualizations of Prompt2Effect's test time adaptation. }
    \label{fig:rotate_right}
\end{figure}

\begin{figure}[t]
    \centering
\includegraphics[width=0.9\linewidth]{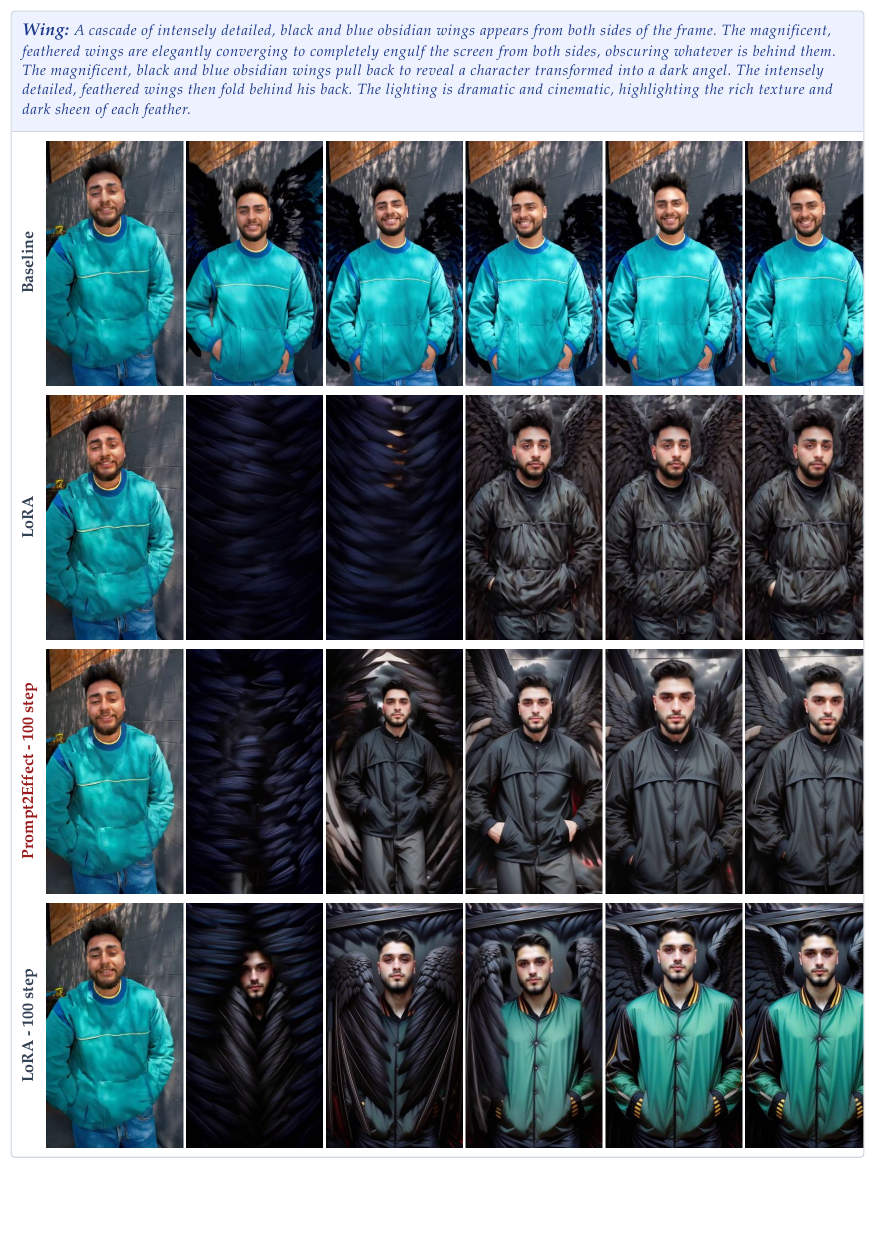}
\vspace{-2em}
\caption{More visualizations of Prompt2Effect's test time adaptation. }
    \label{fig:wing}
\end{figure}

To provide a more comprehensive evaluation of our framework, we present additional qualitative results across both in-distribution and challenging out-of-distribution scenarios. Figure~\ref{fig:iid_viz1} showcases the robust performance of \methodname{} when predicting standard in-distribution effects, demonstrating high visual fidelity and precise alignment with text prompts in a single zero-shot forward pass. Furthermore, to illustrate the efficacy of our test-time adaptation strategy, Figure~\ref{fig:rotate_right} and Figure~\ref{fig:wing} visualize the iterative refinement process on more complex, idiosyncratic motion dynamics. These results highlight how the hypernetwork's initialization allows the model to rapidly converge and capture fine-grained textures and structural transformations with minimal optimization steps.


\section{Limitations and Future Work} \label{sec:limits}
While \methodname{} achieves strong zero-shot performance on in-distribution concepts and demonstrates rapid fast-adaptation capabilities, its zero-shot generalization to highly out-of-distribution (OOD) concepts remains constrained by the scale and diversity of the hypernetwork's training set. Specifically, complex effects involving rare motions, multi-stage sequential transformations, spatially localized edits, or highly ambiguous text prompts may exhibit a concept gap when synthesized directly in a single forward pass. Furthermore, visual elements that fall far outside the distribution of the curated effect library may not perfectly align with user intent in pure zero-shot generation.

To mitigate these limitations in practice, our framework relies on the \textbf{Init-100} refinement mechanism. By leveraging the hypernetwork's predicted weights as a powerful initialization, users can recover fine-grained OOD textures and idiosyncratic details with minimal optimization steps (e.g., 100 steps rather than 1,000). Moving forward, systematically scaling the size and diversity of the target effect LoRA collection during the hypernetwork's pretraining phase should significantly narrow this OOD gap and further improve zero-shot generalization.

Beyond scaling the training data, we identify two promising directions for future research. First, while our current study is rigorously scoped to Image-to-Video (I2V) diffusion models, extending the \methodname{} framework to Text-to-Video (T2V) architectures presents an exciting opportunity to amortize adaptation costs across a broader class of generative video models. Second, incorporating structured, region-aware control into the LoRA prediction process is highly relevant for precise video editing. Future iterations could explore mask-conditioned LoRA synthesis, such as adding spatial or mask tokens to the weight-driven input formulation, to enable localized effect editing and finer region-aware variants, establishing structured LoRA-Edit capabilities.

\end{document}